\documentclass[10pt,journal]{IEEEtran}
\renewcommand{\figurename}{FIGURE}
\usepackage{amsmath,amsfonts}

\usepackage{algorithm} 
\usepackage{algorithmic}  
\usepackage[algo2e]{algorithm2e}

\usepackage{array}
\usepackage[caption=false,font=normalsize,labelfont=sf,textfont=sf]{subfig}

\usepackage{textcomp}
\usepackage{stfloats}
\usepackage{url}
\usepackage{graphicx}
\usepackage{cite}
\usepackage{pifont}
\usepackage{multirow}
\usepackage[switch]{lineno}
\usepackage{enumitem}
\usepackage{threeparttable}
\usepackage{enumerate}
\usepackage{amssymb}
\usepackage{algorithm2e}

\usepackage{float}
\usepackage{setspace}

\usepackage[normal]{caption2}

\usepackage{color}
\usepackage{verbatim}
\usepackage{balance}
\usepackage{soul}
\usepackage{ragged2e}

\usepackage{colortbl}
\usepackage{booktabs}
\usepackage[table]{xcolor}

\definecolor{ACMPurple}{cmyk}{0.55,1,0,0.15}
\definecolor{ACMDarkBlue}{cmyk}{1,0.58,0,0.21}

\usepackage[colorlinks,linkcolor=purple,citecolor=violet,anchorcolor=blue,urlcolor=blue,backref=page]{hyperref}

\newcommand{\cmark}{\ding{51}}%
\newcommand{\xmark}{\ding{55}}%

\usepackage{amsthm}
\newtheorem{theorem}{Theorem}
\newtheorem{definition}{Definition}
\newtheorem{lemma}{Lemma}
\newtheorem{corollary}{Corollary}

\begin{document}


\title{Exploring Open-world Continual Learning with Knowns-Unknowns Knowledge Transfer}

\author{
        Yujie Li,
        Guannan Lai,
        Xin Yang$^*$, \textit{Member, IEEE},
        Yonghao Li,
        Marcello Bonsangue,
        and Tianrui Li, \textit{Senior Member, IEEE}
\thanks{$^*$Xin Yang is the corresponding author (yangxin@swufe.edu.cn).}

\thanks{Yujie Li, Guannan Lai, Xin Yang and Yonghao Li are with the Southwestern University of Finance and Economics, China (E-mail: liyj1201@gmail.com, aignlai@163.com, yangxin@swufe.edu.cn, liyonghao@swufe.edu.cn).}

\thanks{Yujie Li and Marcello Bonsangue are with the Leiden Institute of Advanced Computer Science (LIACS), Leiden University, Netherlands (E-mail: liyj1201@gmail.com, m.m.bonsangue@liacs.leidenuniv.nl).}

\thanks{Tianrui Li is with the School of Computing and Artificial Intelligence, Southwest Jiaotong University, Chengdu, China (e-mail: trli@swjtu.edu.cn).}

\thanks{Manuscript received XX XX, 2025; revised XX XX, 2025.}
}

%
%

\markboth{IEEE TRANSACTIONS ON XXXXX,~Vol.~XX, No.~XX, XX~2025}%
{Shell \MakeLowercase{\textit{Li et al.}}: Exploring Open-world Continual Learning with Knowns-Unknowns Knowledge Transfer}
%





\maketitle
\begin{abstract}
Open-World Continual Learning (OWCL) is a challenging paradigm where models must incrementally learn new knowledge without forgetting while operating under an open-world assumption. This requires handling incomplete training data and recognizing unknown samples during inference. However, existing OWCL methods often treat open detection and continual learning as separate tasks, limiting their ability to integrate open-set detection and incremental classification in OWCL. Moreover, current approaches primarily focus on transferring knowledge from known samples, neglecting the insights derived from unknown/open samples. To address these limitations, we formalize four distinct OWCL scenarios and conduct comprehensive empirical experiments to explore potential challenges in OWCL. Our findings reveal a significant interplay between the open detection of unknowns and incremental classification of knowns, challenging a widely held assumption that unknown detection and known classification are orthogonal processes. Building on our insights, we propose \textbf{HoliTrans} (Holistic Knowns-Unknowns Knowledge Transfer), a novel OWCL framework that integrates nonlinear random projection (NRP) to create a more linearly separable embedding space and distribution-aware prototypes (DAPs) to construct an adaptive knowledge space. Particularly, our HoliTrans effectively supports knowledge transfer for both known and unknown samples while dynamically updating representations of open samples during OWCL. Extensive experiments across various OWCL scenarios demonstrate that HoliTrans outperforms 22 competitive baselines, bridging the gap between OWCL theory and practice and providing a robust, scalable framework for advancing open-world learning paradigms.
\end{abstract}



\section{Introduction}
\label{Introduction}
\begin{figure}[ht]
    \centering
\includegraphics[width=\columnwidth]{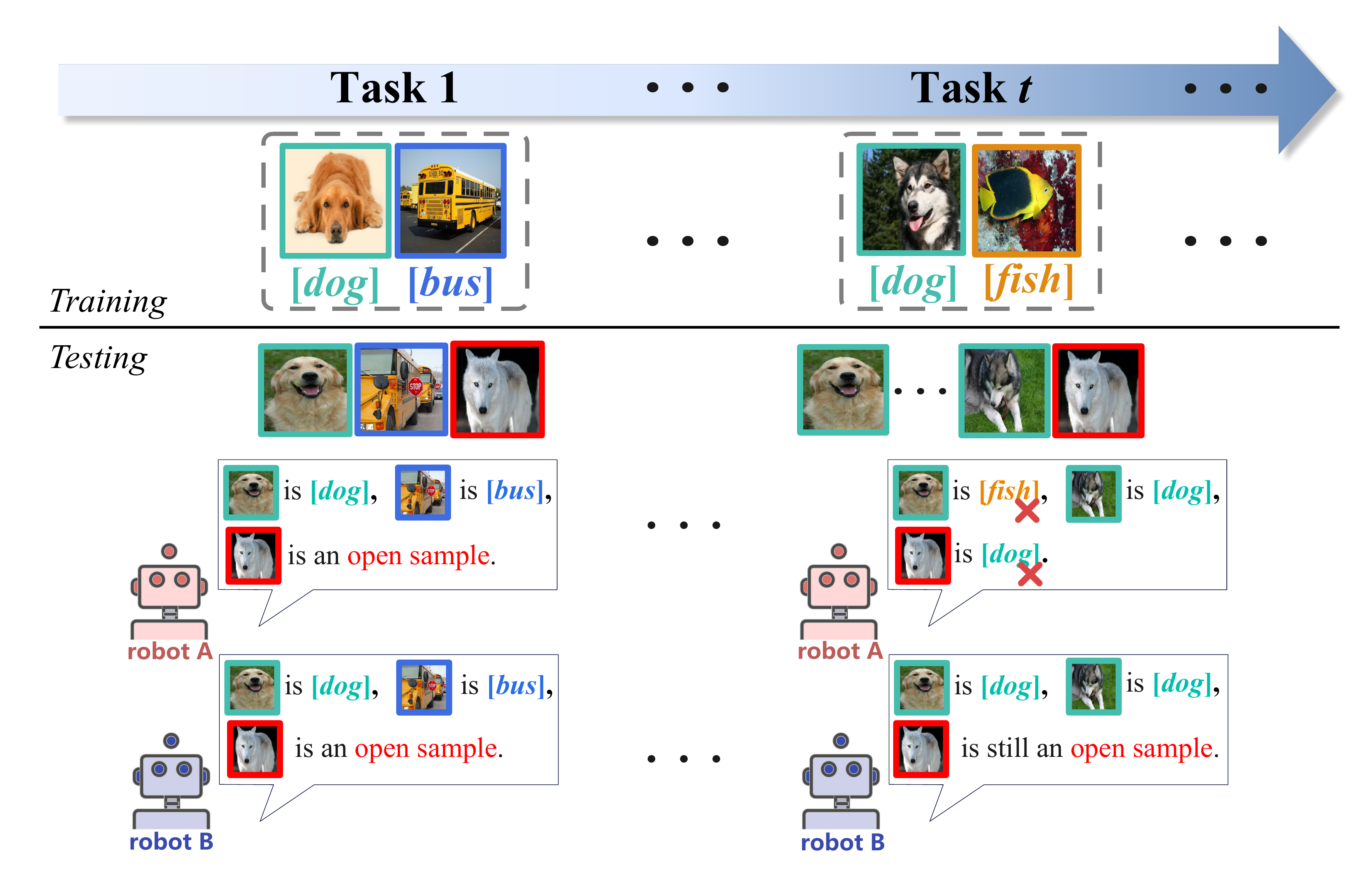}
    \caption{\textbf{A toy example of the KIRO scenario in OWCL.} 
During training Task 1 and Task t, Robots A and B learn \textit{Golden Retrievers} and \textit{Labradors}, both labeled as [\textit{dog}]. In testing Task 1, both robots correctly classify [\textit{dog}] and [\textit{bus}] samples while identifying the unknown wolf without errors. However, in testing Task 2, Robot A struggles with Golden Retrievers and misclassifies the wolf as [\textit{dog}] due to its similarity to Labradors. In contrast, Robot B accurately classifies all known samples and continually detects the wolf as an open sample with knowledge transfer.}
    \label{fig:KIRO illustration}
\end{figure}

Open-World Continual Learning (OWCL) \cite{kim2022theoretical, liu2023ai} represents a highly practical yet profoundly challenging machine learning paradigm. In OWCL, a model must continually adapt to an unbounded sequence of tasks in a dynamic open environment \cite{scheirer2012toward, bendale2015towards}, where novelties might emerge in testing unpredictably over time \cite{kirkpatrick2017overcoming, pmlr-v70-zenke17a, wang2024comprehensive}.
Unlike traditional learning models that operate in a closed and fixed set of classes, OWCL aims at learning on the job in an open-world assumption with the goal of recognizing unseen/open samples and incrementally acquiring knowledge from new tasks without forgetting \cite{kishida2021object,truong2023fairness,kim2025open}.
Due to the potential occurrence of novelties in continual learning (CL) \cite{zhou2024class}, OWCL models need to accurately detect unknowns to prevent unknown samples from being incorrectly classified into known categories. At the same time, OWCL requires the model to retain previously learned knowledge without forgetting while continually performing open detection.
In summary, the open detection for unknowns and classification for knowns are \textit{interdependent} in OWCL: on the one hand, the presence of unknowns complicates the trade-off between \textit{knowledge stability} \cite{parisi2019continual,zhou2024class} and \textit{knowledge plasticity} \cite{dhar2019learning,vijayan2023trire}; on the other hand, the incremental learning of new tasks makes open detection in an embedding space more challenging with the expanding knowledge space.

Despite growing attention to OWCL in recent years, current approaches \cite{mundt2022unified, kim2025open} still treat OWCL as a simple combination of open-set recognition and CL, rather than as an integrated paradigm, making it only effective in knowledge transfer related to known samples, while neglecting the knowledge derived from unknown samples. 
Therefore, a promising OWCL model must be capable of \textbf{knowns-unknowns knowledge transfer}, i.e., effectively transferring knowledge both for known categories and unknown samples.
Besides, there is a lack of comprehensive problem formulation and thorough empirical explorations of potential issues in OWCL, making it difficult to compare the performance of existing methods and making it unclear how to choose one method over another.
In response, this paper explores the issues arising in OWCL and adopts an integrative perspective to deal jointly with unknown samples' detection and known samples' classification, particularly the knowledge transfer for both unknowns and knowns.

\begin{table*}[ht]
\centering
\caption{Overview of the Four OWCL Scenarios.}\label{tab:scenarios}
\vspace{2mm}
\begin{tabular}{ccccc}
\toprule
\textit{\textbf{Scenario}} & \textbf{CINO} & \textbf{CIRO} & \textbf{KINO} & \textbf{KIRO} \\ \midrule
Known Classes Repeatedly Appear & \xmark  & \xmark  & \cmark  & \cmark  \\ \midrule
Unknown Samples Repeatedly Appear & \xmark & \cmark  & \xmark  &  \cmark \\ \bottomrule
\end{tabular}
\end{table*}

\begin{figure*}[ht]
\centering
\includegraphics[width=\textwidth]{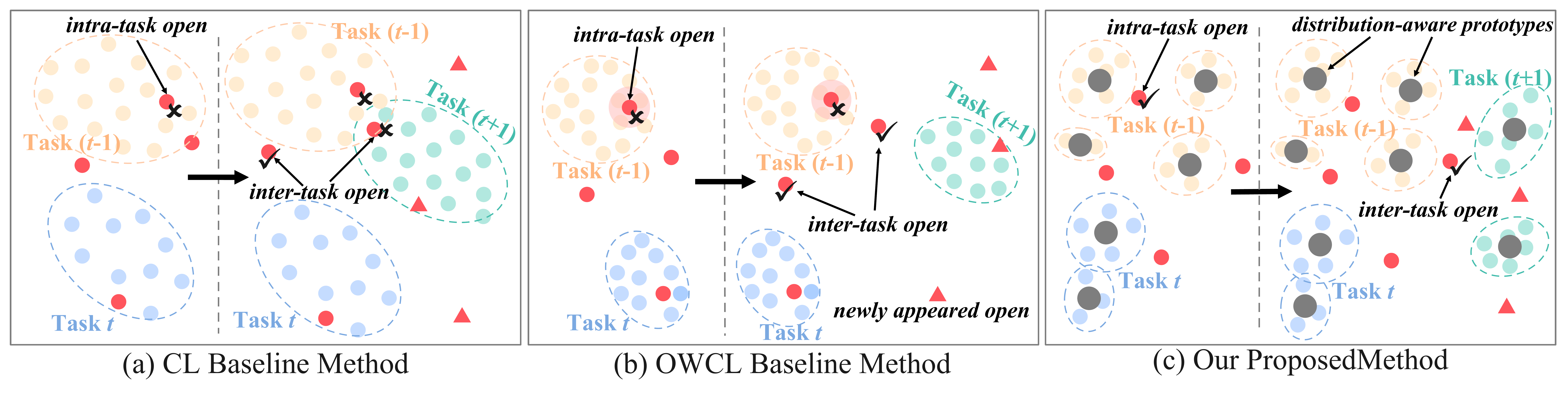}
    \caption{From task \( t \) to task \( t+1 \) in the KIRO scenario, we illustrate the embedding space utilizing different baselines. Red circles and red triangles denote unknown samples encountered in tasks \( t \) and \( t-1 \), respectively. Orange, blue, and green circles represent the labeled training data learned during tasks \( t-1 \), \( t \), and \( t+1 \), respectively. Dashed lines depict the decision boundaries inferred by the model during OWCL. \textbf{(a)} The original embedding distributions for each task learned by a classical CL baseline, EWC. \textbf{(b)} A current competitive OWCL approach enforces tighter embedding cohesion for each task. \textbf{(c)} Our proposed method concurrently addresses open detection and incrementally reduces known sample classification errors with knowledge transfer. [Best view in color]}
    \label{fig:intro}
\end{figure*}

As shown in \autoref{fig:KIRO illustration}, in real-world scenarios, labeled samples for a new category are rarely available all at once or from a single task \cite{truong2023fairness}; instead, they are acquired incrementally across different tasks over time \cite{masana2022class, lo2022adversarially}.
Hence, in OWCL, a certain category may repeatedly appear across different tasks with varying data distributions.
Moreover, as new tasks are learned incrementally, the boundaries between known categories become increasingly ambiguous.
As a result, the repeated appearance of open/unknown samples further makes it difficult for the OWCL model to distinguish between the unknown and known samples.
Accordingly, as outlined in \autoref{tab:scenarios}, there are two fundamental principles for refining OWCL scenarios: \textit{(1) whether a known class with changing data distribution appears in different tasks during the training phase, and (2) whether an unknown sample repeatedly appears in different tasks during the testing phase.} 

Therefore, based on the principles of scenario categorization, we extend existing CL scenarios (i.e., task-incremental learning, domain-incremental learning, and class-incremental learning) to the context of OWCL.
However, different from CL, OWCL does not permit task identification, as unknown samples may randomly appear during testing \cite{li2024learning}. 
Thus, task-incremental and domain-incremental learning are insufficient for describing the OWCL scenarios. 
In response, we introduce a novel scenario definition, termed knowledge-incremental learning (\textbf{KIL}): \textit{models must be able to solve each task seen so far without knowing which task is being performed, where the training sets across different tasks are not strictly non-overlapping; that is, shifts in the distribution of known categories may occur.}
KIL incorporates class-incremental learning and accommodates shifts in the distribution of specific categories, thereby providing a more realistic scenario description for OWCL and posing greater difficulty.
Accordingly, we then introduce four distinct scenarios for OWCL of increasing difficulty: class-incremental with non-repetitive open samples (\textbf{CINO}), class-incremental learning with repetitive open samples (\textbf{CIRO}), knowledge-incremental learning with non-repetitive open samples (\textbf{KINO}), knowledge-incremental learning with repetitive open samples (\textbf{KIRO}).

Recently, much research \cite{schlachter2020deep,zhu2022multi,huang2022class} highlighted the significance of both inter-task and intra-task open samples in developing effective open-set recognition. 
Hence, this work proposes an OWCL model to distinguish inter-task open samples from each previously learned task, where inter-task open samples randomly appear across different tasks.
Meanwhile, the OWCL model must differentiate intra-task open samples from known categories within the same task while preserving accurate classifications of the known categories.
Then, to explore potential issues of OWCL, we conduct extensive comparisons with current competitive baselines under the most challenging scenario, KIRO.

From the experimental results, we observed that continually detecting novelties in OWCL requires consideration of inter-task and intra-task open samples. Moreover, there is an intrinsic interplay between open detection and classification for known categories.
Specifically, as illustrated in \autoref{fig:intro} (a), we employ a classical CL baseline, EWC \cite{kirkpatrick2017overcoming}, to learn the original embedding space under the KIRO scenario. It can be observed that with the incremental acquisition of new tasks and the recurring appearance of open samples, the embedding distributions of different tasks may overlap, rendering the model incapable of detecting inter-task open samples.
Moreover, due to the similarity between certain open samples and existing known classes, some intra-task open samples might be erroneously classified into known categories.
In \autoref{fig:intro} (b), the competitive OWCL baseline Pro-KT \cite{li2024learning} improves task boundaries for detecting inter-task unknown samples but adversely impacts the detection of intra-task open samples. This occurs because intra-task open samples within the convex hull \cite{zhou2022knn} of known categories (as denoted by pink shadows) are prone to misclassification as known categories.
Therefore, as evidenced by \autoref{fig:intro} (a) and (b),  both inter-task and intra-task open samples significantly amplify the open risk associated with unknowns within the unified knowledge space, further exacerbating incremental classification errors of knowns.

Our experimental findings reveal a significant interaction between open detection and incremental prediction in OWCL, while existing OWCL works \cite{kim2022theoretical,truong2023fairness,kim2025open} claimed that the open detection and the incremental classification of known samples are orthogonal.
Thus, the existing OWCL methods rely on a simple continual learning framework combined with an open detection module, which are only effective in the CINO scenario but fail in more complex OWCL settings.
This motivates the need for models to enable effective knowledge transfer—not only to handle both known and unknown samples but also to update and expand the knowledge space for OWCL incrementally.
Accordingly, we establish a unified formulation to address the OWCL problem, grounded in robust theoretical principles.
To tackle the challenge of knowns-unknowns knowledge transfer, we introduce the \textbf{nonlinear random projection (NRP)} for OWCL to facilitate the learning of a more linearly separable embedding space, thereby reducing errors in knowledge transfer for known samples.
Meanwhile, we propose a \textbf{distribution-aware prototypes (DAPs)} approach, leveraging generative replay with novel pseudo-open samples to transfer knowledge related to novelties while updating the representation of open samples during the OWCL process.

In summary, this work presents a novel framework for transferring knowledge of both known and unknown samples in OWCL, termed \textbf{HoliTrans} (holistic knowns-unknowns knowledge transfer), designed to address all OWCL scenarios comprehensively and bridge the theory-practice gap.
The proposed HoliTrans is a general framework with solid theoretical support that can be easily applied to any OWCL scenario. In HoliTrans, we present nonlinear random projection (NRP) with distribution-aware prototypes (DAPs) that construct an adaptive knowledge space, support knowledge transfer for known and unknown samples and even update knowledge during learning new tasks with novelties.
The contributions of this work are:
\begin{itemize}[leftmargin=*]
    \item This work presents a formal construction for OWCL, delineating four distinct scenarios. Empirical experiments find a significant interaction between open detection and incremental prediction in OWCL, challenging the current assumption of orthogonal decomposition.
    \item This paper introduces a novel OWCL framework, HoliTrans, which transfers knowledge from both known and unknown samples, and thus achieves the open detection and incremental prediction quality in a unified manner.
    \item Extensive experiments validate our theoretical findings, with HoliTrans outperforming 22 competitive baselines across different OWCL scenarios, and our code is open-sourced on https://github.com/AIGNLAI/HoliTrans.
\end{itemize}

\section{Related work}
\label{Related work}
Continual learning (CL) \cite{kirkpatrick2017overcoming,li2017learning} usually operates under the closed-world assumption, where the system assumes that all test or deployment samples belong to one of the predefined classes seen during training \cite{bendale2015towards,fei2016learning,liu2024task}.
However, this assumption implies no exposure to novel or previously unseen data during testing, which is far from realistic in dynamic, real-world environments \cite{liu2023ai,li2024learning,liu2023ai}. 
In practice, continual learning systems are frequently confronted with new, unknown classes, necessitating the ability to detect, adapt to, and incrementally learn these novelties, continually acquiring new knowledge over time \cite{wang2020novelty,yu2022self,aljundi2022continual}.
Hence, it is imperative to detect and incrementally learn novelties while acquiring knowledge without forgetting over time.

More recently, continual learning in an open world or simply \textit{Open-world Continual Learning} (OWCL) has been appealing yet challenging with increasing works \cite{zhou2024class,liu2023ai}. 
In order to enable existing CL models to effectively detect open/unknown samples, primary OWCL research utilized open-set recognition (OSR) and out-of-distribution (OOD) detection methods as expansive components into CL baselines to tackle the OWCL tasks. 
\cite{8631004-2019} proposed an OSR framework based on extreme value theory, incorporating incremental tasks to manage dynamic learning environments. \cite{Joseph_2021_CVPR} introduced an approach leveraging contrastive clustering and an energy-based identification method \cite{chan2021entropy} for handling dynamic data, enabling the system to recognize and accommodate novel inputs during continual learning.
Building on these previous efforts, recent OWCL studies emphasized the integration of OOD detection techniques within the continual learning paradigm. For instance, \cite{kim2022theoretical,li2024learning,kim2025open} highlighted the importance of novelty detection as a crucial aspect of open-world learning, suggesting that existing OOD techniques could be effectively adapted to the continual learning setting. 
In a complementary development, frameworks such as SOLA \cite{liu2023ai,mazumder2024lifelong} have been proposed, combining OOD detection with incremental task adaptation to facilitate novelty detection and task-specific learning in an open-world context. 

Nevertheless, current research still relies on simplistic approaches by combining CL methods with OOD detection components \cite{wang2020novelty,schlachter2020deep,aljundi2022continual,mundt2022unified,li2024learning,kim2025open}, and several crucial challenges persist in OWCL. 
First, the absence of a standardized and general problem formulation makes it challenging to compare the performance of existing methods in a consistent setting, leading to fairness issues. 
Second, there are still experimental and theoretical gaps in exploring the knowledge transfer for known and unknown samples in OWCL.
Finally, there is a lack of theoretical foundation to guide the design of an OWCL model that supports knowledge transferring and knowledge updating.

To address these limitations, this paper makes several key contributions. 
First, we provide rigorous theoretical analyses with a formal problem construction for understanding the variations and challenges inherent to OWCL. 
By constructing four distinct scenarios, we conduct extensive empirical experiments and systematically explore the factors influencing OWCL model performance under these different scenarios, identifying a significant interplay between open detection and the classification of known samples. 
Finally, we propose a novel framework, termed HoliTrans, designed to effectively address knowns-unknowns knowledge transfer by introducing NRP and proposing DAPs in an adaptive knowledge space.
The proposed HoliTrans serves as a strong baseline for future research, demonstrating its robustness and efficacy across a range of OWCL scenarios with abundant experiments.

\section{Understanding OWCL: Formulation and Preliminary Experiments}
\label{Preliminary}
In this section, we first formalize the problem of OWCL by integrating the optimization with open risk from unknowns and incremental prediction errors from knowns.
Then, we provide an in-depth categorization scheme for different OWCL scenarios.
Subsequently, to explore potential issues in OWCL, we conduct abundant experiments with baselines using different kinds of classifiers (i.e., Softmax-based and NCM (Nearest Class Mean)-based classifiers) on different baselines.

\subsection{Problem Formulation}
In OWCL, samples not seen in the training set may appear during testing, thus, we use $\mathcal{D}_{tr}$ and $\mathcal{D}_{te}$ to distinguish between the training set and test set; all superscripts indicate the task order. 
We list the notations throughout this paper in \autoref{tab: notations} for better clarification.

\begin{table}[ht]\small
 \begin{center}
\caption{Notations and explanations.} \label{tab: notations}
\vspace{2mm}
  \begin{tabular}{c c}
    \toprule
    \textbf{Notation} & \textbf{Explanation} \\ 
    \hline
    $t,i$ & Task identifiers \\
    $h^t$ & Neural network after task $t$\\
    $D^i_{tr}$ & Training set of task $i$\\
    $D^i_{te}$ & Testing set of task $i$\\
    $N_i$ &  Number of training samples in $D^i_{tr}$\\
    $M_i$ &  Number of seen classes in $D^i_{tr}$\\
    $\mu(h(x), h(D^i_{tr}))$ & Open risk of $x$ in $D^i_{te}$\\
    $\mathcal{H}$ & Universal hypothesis space\\
    $\epsilon_{\mathcal{D}_{tr}^{i}}(h)$ & General prediction error on task $i$\\
    $\lambda$ & Balance for open risk and classification\\
    $\mathcal{S}$ & Universal space\\
    $\mathcal{O}$ & Open space for OWCL\\
    $\mathcal{K}$ & Known space from each trained task \\
    $\widehat{\epsilon}_{\mathcal{D}_{tr}^{i}}(h^t)$ & Empirical prediction error \\
    $\mathcal{B}^{t}$ & Experience buffers from $\mathcal{D}_{tr}^{t}$ \\
    $\boldsymbol{s_x} $ & Max similarity of $x$ to prototypes \\
    $g(\cdot)$ & Nonlinear function \\
    $\phi(\cdot)$ & Pre-trained model\\
    $\mathcal{P}$ & Projection layer\\
    $G^i$ & Gram matrix of NRP feature \\
    $\boldsymbol{p}_{k}^i$ & Prototype of class $k$ in task $i$\\
    $\mathcal{D'}_{te}^i$ & New test set with pseudo-samples \\
    \bottomrule
  \end{tabular}
 \end{center}
\end{table}

Currently, there is a lack of consensus regarding the OWCL problem formulation within the community. In response, we develop a comprehensive and general problem definition, taking into account both the inherent open risk and the incremental classification error that arise in all OWCL scenarios as follows:
\begin{definition}\label{def:probelm}
    \textbf{(OWCL Problem Formulation.) }
    Each task $i$ has $N_i$ training samples and $M_i$ classes $\mathcal{D}^{i}_{tr}=\{ \boldsymbol{x}_j^{i} \in \mathcal{X}_{j=1}^{N_i},y_j^{i} \in \mathcal{Y}_{j=1}^{M_i}\}$. 
In the training phase, only data pertinent to the current task is accessible, while the test samples $\mathcal{D}^{i}_{te}=\{ \boldsymbol{x}_j^{i} \in \mathcal{X}_{j=1}^{N'_{i}},y_j^{i} \in \mathcal{Y}_{j=1}^{M'_{i}}\}$ may contain unknowns/novelties, i.e., $\mathcal{Y}_{j=1}^{M'_{i}} - \mathcal{Y}_{j=1}^{M_i} \neq \emptyset $. 
Given the latent representation of the current task $t$'s training set $h(\mathcal{D}^{t}_{tr})$, we denote $\mu(h(\boldsymbol{x}),h(\mathcal{D}_{tr}^{i}))$ as the open risk of a sample $\boldsymbol{x}$ from task $i$'s test set $\mathcal{D}^{i}_{te}$.
The goal of Open-world continual learning (\textbf{OWCL}) is to learn a uniform function $h^*$ that minimizes the \textbf{open risk} and \textbf{incremental prediction error} jointly, across all seen tasks $[1,t]$:
\begin{equation}\label{eq:problem}
\arg \min _{h^* \in \mathcal{H}} \{ (1-\lambda) \underbrace{\sum_{i=1}^{t} \mu(h(\boldsymbol{x}),h(\mathcal{D}_{tr}^{i}))} _{\textbf{(A)}} + \lambda \underbrace{\sum_{i=1}^{t} \epsilon_{\mathcal{D}_{tr}^{i}}(h)}_{\textbf{(B)}} 
 \},
\end{equation}
where $\mathcal{H}$ is a universal function space, $\epsilon_{\mathcal{D}_{tr}^{(i)}}(h)$ is the generally suggested prediction error term on task $i$ \cite{bendale2015towards}, $\mu$ can be defined on any scoring function for OOD detection, $\lambda$ is a hyper-parameter to balance the open risk and the incremental prediction error.
\end{definition}

\subsection{Four OWCL Scenarios}\label{sec3.1}

As discussed previously, traditional CL is usually categorized into three scenarios, i.e., \textit{incremental learning, domain-incremental learning and class-incremental learning}, based on whether task identifications are provided during tests \cite{van2018three}. 
However, in OWCL, \textbf{it is impossible for the model to know the task identifiers, because unknown/open samples may randomly appear alongside known ones during testing.} 
Furthermore, \textbf{the shifting distributions of known classes and the repetitive presence of open samples hinder the current CL three scenarios \cite{van2018three} from being compatible with OWCL problems.}
Consequently, four distinct scenarios of increasing difficulty (presented in \autoref{tab:scenarios}).

First, in the class-incremental learning with non-repetitive open samples (\textbf{CINO}) scenario, the training set for each task is introduced incrementally without repeated classes. Once previously unseen samples are encountered during testing, they do not reappear in future tasks. This scenario is ideal for applications that require continual learning of new classes without retaining historical data, such as one-time classification tasks in fields like species identification or material recognition.

In the second scenario, class-incremental learning with repetitive open samples (\textbf{CIRO}), the training set also grows incrementally without repeated classes, but open samples encountered in previous tasks may reoccur in future tests. This scenario is particularly suited to dynamic environments, such as autonomous driving or robotic systems, where the system must handle repetitive open samples over time, such as environmental objects like traffic signs or obstacles.

The third scenario, knowledge-incremental learning with non-repetitive open samples (\textbf{KINO}), involves training sets that evolve incrementally, with repeated classes and shifting distributions, but open samples encountered during testing do not reappear. This approach is applicable to domains that need to adapt to changing class distributions without recalling past open samples, such as streaming data analysis or online content recommendation systems, where rapid adaptation to evolving user preferences is crucial.

Finally, in knowledge-incremental learning with repetitive open samples (\textbf{KIRO}), the training set evolves incrementally with repeated classes and changing distributions, while previously unseen open samples encountered during testing may reoccur. This scenario is most relevant for complex, dynamic real-world applications, such as intelligent surveillance systems or financial risk analysis, where models need to learn from historical patterns and manage repetitive open samples, such as shifts in customer behavior or market trends.

\textbf{\textit{Remarks.}} Knowledge-incremental learning poses greater difficulty than task-, domain- and class-incremental learning settings. 
The primary difference is that knowledge-incremental learning emphasizes adapting to potential shifts in the data distributions of previously learned tasks while incrementally learning new classes without task identifiers. 
Due to the presence of open samples and complex task distribution shifts in OWCL, along with the inability to know task identifiers during testing, knowledge-incremental learning requires the model to transfer knowledge not only about known samples but also about unknown samples.

\subsection{Empirical Experiments on All OWCL Scenarios}\label{sec3.2}
In this subsection, we begin by thoroughly examining potential issues within OWCL. Most excitingly, many of our empirical results challenge prevailing claims in current OWCL research, questioning existing assumptions within the field. Our findings further inspire us to explore related theories and propose a novel OWCL model accordingly.

Recently, with the increasing application of pre-trained models and growing research on Nearest Class Mean (NCM) classifiers, current CL studies \cite{mcdonnell2024ranpac} found that the NCM strategy has yet to reach its full potential for accuracy. Remarkably, it can achieve standout performance when combined with carefully tailored strategies that enhance feature representations extracted from pre-trained models.
Hence, given the distinct characteristics and performance exhibited by different classifiers and loss calculation methods, we conduct experiments using various approaches with different types of classifiers (i.e., Softmax-based and NCM-based) across four OWCL scenarios.
Specifically, as typical methods utilizing a Softmax-based classifier, fine-tuning and L2P \cite{wang2022learning} are current state-of-the-art (SOTA) approaches for handling continual learning (CL). SimpleCIL \cite{zhou2023revisiting} and RanPAC \cite{mcdonnell2024ranpac} represent two of the latest and most competitive continual learning baselines employing an NCM-based classifier.

\begin{figure}[ht]
    \centering
    \includegraphics[width=\columnwidth]{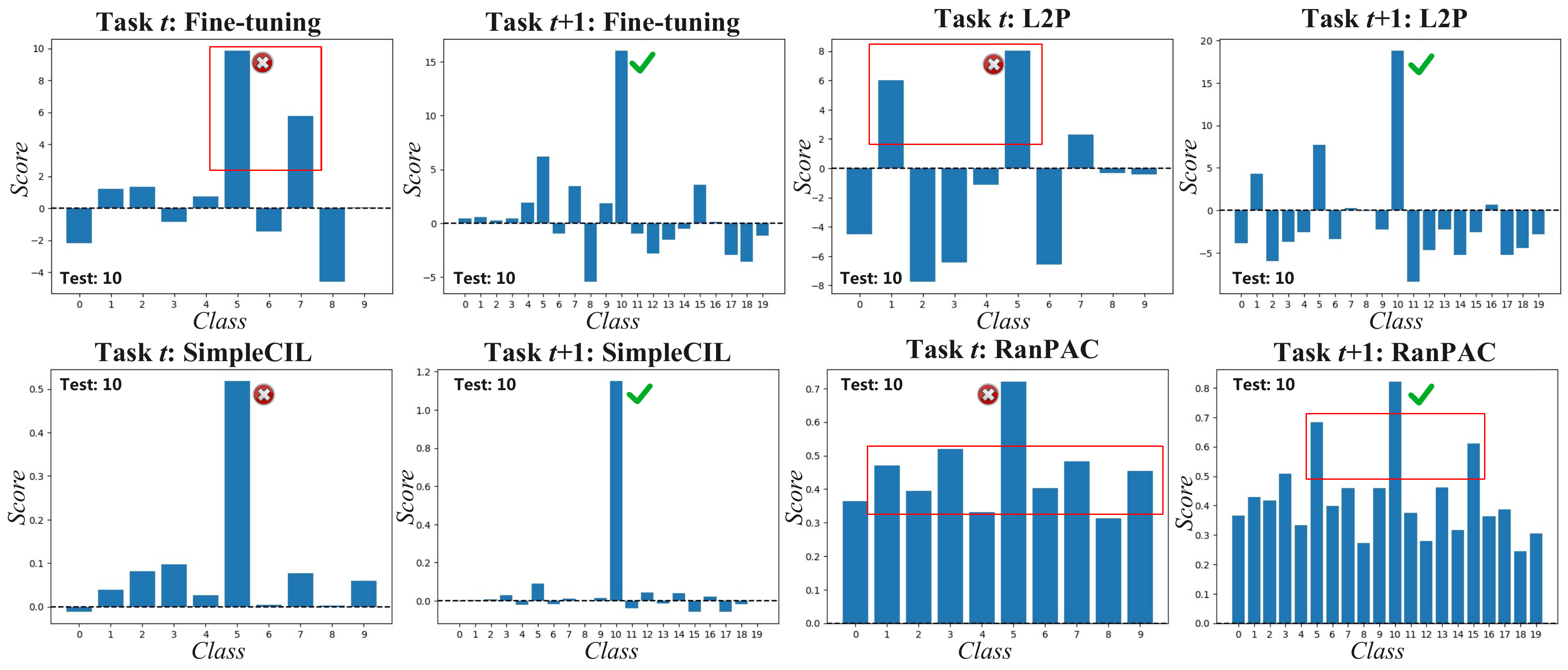}
    \caption{Unscaled Classification Scores on the CINO scenario.}\label{CINO}
\end{figure}

\begin{figure}[ht]
    \centering
    \includegraphics[width=\columnwidth]{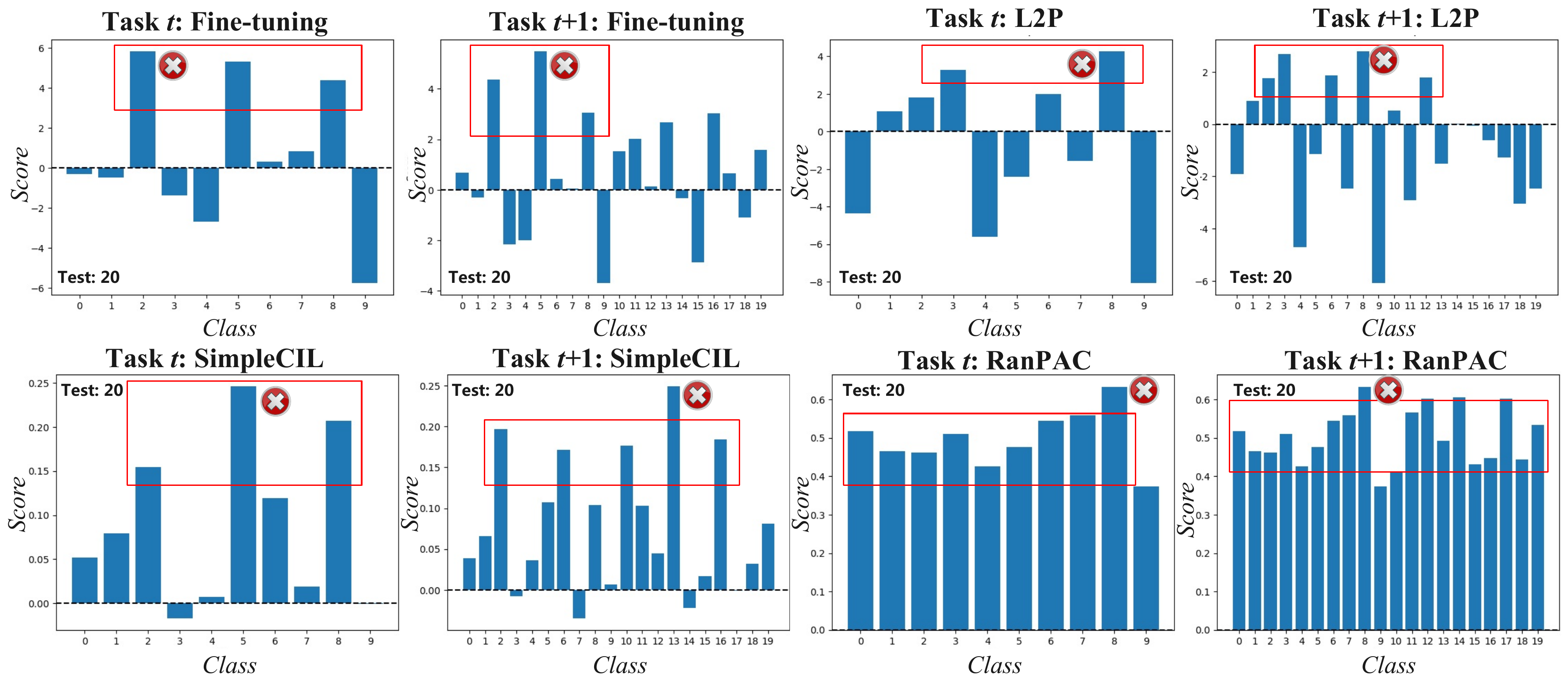}
    \caption{Unscaled Classification Scores on the CIRO scenario.}\label{CIRO}
\end{figure}

\begin{figure}[ht]
    \centering
    \includegraphics[width=\columnwidth]{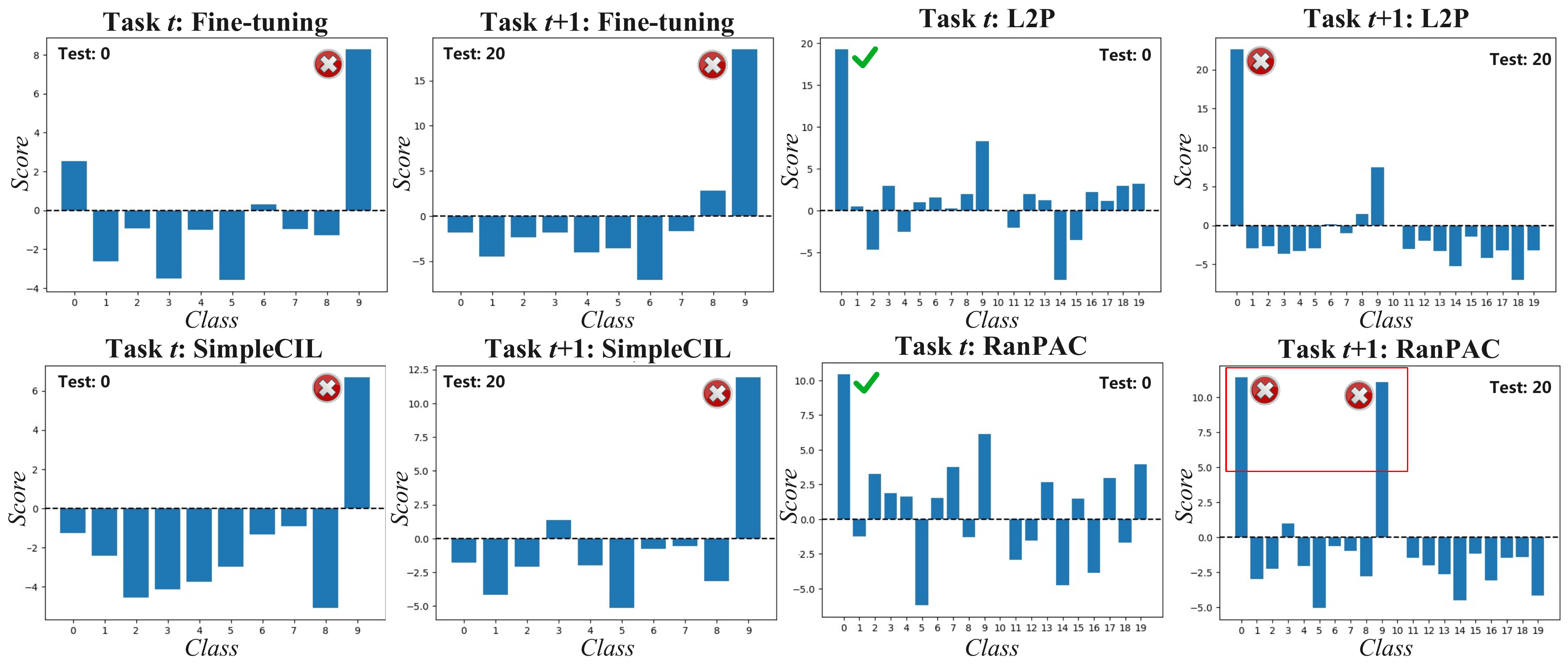}
    \caption{Unscaled Classification Scores on the KINO Scenario.}\label{KINO}
\end{figure}

\begin{figure}[ht]
    \centering
    \includegraphics[width=\columnwidth]{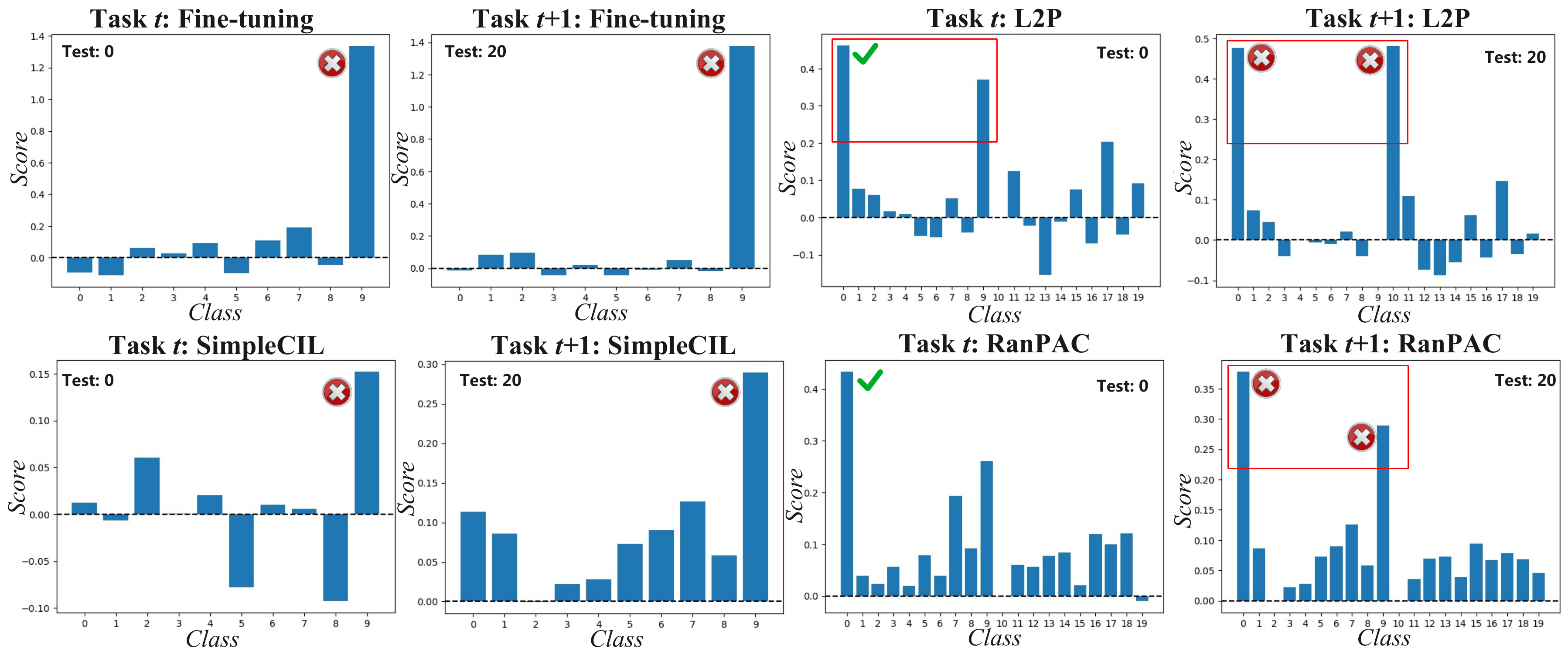}
    \caption{Unscaled Classification Scores on the KIRO Scenario.}\label{KIRO}
\end{figure}

\autoref{CINO} to \autoref{KIRO} respectively display the performance of different models across four distinct OWCL scenarios, with the horizontal axis representing class labels and the vertical axis indicating unnormalized classification scores calculated by different types of classifiers.

From \autoref{CINO}, we observe that due to the CINO scenario, each task's training set is presented incrementally without repeated classes, and previously unseen samples encountered during testing do not reappear subsequently. Therefore, open-world continual learning models fail to detect open samples appearing in task $t$, but after learning the corresponding training data for these open samples in task $t+1$, the models are then able to classify them.

From \autoref{CIRO}, in the CIRO scenario, because open samples encountered previously may reoccur in subsequent tests, we find that all the baseline models fail to detect open samples in both task $t$  and task $t+1$, erroneously classifying the inter-task and intra-task open samples into known classes, which results in significant open risk. Additionally, we notice that in the CIRO scenario, the unnormalized classification scores for each class are very close, reflecting the complexity of the data distributions among various tasks. By comparing the results in CINO and CIRO, we conclude an observation: existing continual learning and incremental learning methods cannot validate either inter-task nor intra-task open samples and tend to classify open samples into seen categories erroneously. 

Next, in the KINO and KIRO scenarios, each task's training set evolves incrementally with repeated classes and changing distributions. Hence, from \autoref{KINO} and \autoref{KIRO}, it is observed that the models fail to effectively distinguish open samples in task $(t+1)$. Moreover, due to the presence of recurrent open samples in the KIRO scenario, the classification scores obtained by the models in this scenario are also closer. Consequently, another observation can be drawn: the variation in data distribution between tasks poses greater challenges to OWCL, and there exists a mutual influence between open risk and incremental prediction quality.

In conclusion, by comparing all the results from \autoref{CINO} to \autoref{KIRO}, as the difficulty of the four scenarios increases, the classification performance on known classes decreases because known classes repeatedly appear with changing data distributions. Moreover, we find that the NCM-based classifier methods performed better than the Softmax-based classifier methods, especially when the data distributions continuously changed across new tasks and open samples reappeared over time.

\subsubsection{Experimental Findings Summary}
The empirical experiments show that in all OWCL scenarios, the NCM-based classifier methods performed better than the Softmax-based classifier methods, especially when the data distributions changed across new tasks and open samples reappeared over time.
In addition, we conclude with three key findings:
(1) Vanilla continual learning and incremental learning methods are unable to validate open samples, either between tasks or within tasks, and tend to classify these open samples into previously seen categories mistakenly.
(2) As the difficulty of the four scenarios increases, the classification performance on known samples decreases because known classes repeatedly appear with changing task distributions.
(3) Open risk and incremental prediction error coexist in an interplay rather than independently, making them inseparable through orthogonal decomposition.

Our empirical observations demonstrate that the assumptions made in previous works \cite{kim2022theoretical,truong2023fairness} —specifically, that the decomposition of open risk and incremental prediction error is orthogonal and unaffected by each other—are invalid. 
Our findings highlight the limitations of current OWCL approaches and strongly motivate us to design a comprehensive problem formulation with theoretical bases for OWCL in \autoref{theory}. 
Therefore, as open risk and known classification errors cannot be orthogonally decomposed, OWCL models not only transfer knowledge from known samples but also obtain and transfer knowledge from unknown samples.
Accordingly, in \autoref{Method}, we propose a novel framework, HoliTrans, which incorporates a nonlinear random projection method and distribution-aware prototypes to enable the knowledge transfer of both known and unknown samples (termed \textbf{knowns-unknowns knowledge transfer}).

\section{Theoretical Bases}
\label{theory}
\subsection{Open Risk and Incremental Prediction Error}
This section defines a new knowledge space and introduces a generic incremental prediction error term for OWCL. We theoretically prove a tighter bound for incremental prediction error, providing several insights for designing better OWCL algorithms by considering knowledge from knowns and unknowns.

Following a general definition from\cite{bendale2015towards}, we extend the definition of \textbf{open space} $\mathcal{O}$ for OWCL as follow:
\begin{definition}\label{open space}
    \textbf{(Open Space for OWCL.) }
    \begin{equation}
        \mathcal{O} = \mathcal{S}-\bigcup_{\boldsymbol{x} \in \left\{ \mathcal{D}_{tr}^i \right\}_{i=1}^{t}}\mathcal{K}(\boldsymbol{x}),
    \end{equation}
where $\mathcal{S}$ is the universal space, and $\mathcal{K}$ is the known space obtained by each trained task $i \in [1,t]$. 
\end{definition}
Note that it is not necessary to classify each open sample into exactly true unknown classes. 
For the sake of simplicity, we follow the general implementations \cite{fang2021learning}, setting all unknown samples to one unified unknown as $[un]$. 
Hence, given the problem formulation in \autoref{def:probelm}, we assume that $\mathcal{Y}_{j=1}^{M'_{i}} - \mathcal{Y}_{j=1}^{M_i} = \left\{y_{un}\right\}$ where $y_{un}$ represents the one unknown class and $M'_{i}=M_i+1$. Therefore, from the probabilistic perspective, we can refine the left summand term (A) in \autoref{eq:problem} as:
\begin{theorem}\label{the:open risk}
    \textbf{(Open Risk of OWCL.) }
    \begin{equation}
    \begin{aligned}
    & \sum_{i=1}^{t} \mu(h(\boldsymbol{x}),h(\mathcal{D}_{tr}^{i}))  
    = R_{\mathbf{P}, un}(h) \\
    &\ := \int_{\sum_{i=1}^{t} \mathcal{X}_{j=1}^{N'_i}} \ell\left(h(\boldsymbol{x}), y_{un}\right) \mathrm{d} \mathbf{P}_{X \mid Y=y_{un}}(\boldsymbol{x}),
    \end{aligned}
    \end{equation}
    where $\ell$ is a loss satisfying $\ell(y,y')=0 \text{ iff } y=y'$. 
\end{theorem}
Previous works \cite{kim2022theoretical,truong2023fairness} claimed that the decomposition in \autoref{eq:problem} is orthogonal, meaning that the terms (A) and (B) do not affect each other. 
However, from observations in \autoref{sec3.2}, we find that the open risk affects the incremental prediction quality, and that the boundary across the learned tasks also affects the open detection performance.
Therefore, we need to consider the interaction of open risk (term (A) in \autoref{def:probelm}) and incremental prediction error (term (B) in \autoref{def:probelm}) in a unified manner.

Subsequently, we introduce a novel incremental prediction error term and theoretically derive a more compact bound for OWCL:

\begin{theorem}\label{the:task specific error}
\textbf{(Task-Specific Prediction Error.) }After training a new task $t$, the empirical prediction error $\widehat{\epsilon}_{\mathcal{D}_{tr}^{i}}(h^t)$ of the current model $h^t$'s
for an arbitrary task $i (i \leq t)$ is defined as \cite{shi2024unified}:
\begin{equation}
\widehat{\epsilon}_{\mathcal{D}_{tr}^{(i)}}(h^t) = \left\{\begin{array}{ll}
\frac{1}{N_{i}} \sum_{\boldsymbol{x} \in \mathcal{X}_{j=1}^{N_i}} \mathbb{I}({h^t(\boldsymbol{x}) \neq g_{i}(\boldsymbol{x})}) & \text { if }\  i = t, \\
\frac{1}{\widetilde{N}_{i}} \sum_{\boldsymbol{x} \in \widetilde{\mathcal{X}}_{j=1}^{\widetilde{N}_i}} \mathbb{I}({h^t(\boldsymbol{x}) \neq g_{i}(\boldsymbol{x})}) & \text { if }\  i<t .
\end{array}\right.
\end{equation}
$\mathbb{I}(\cdot)$ is an indicator function. Only a small subset of data from previous tasks $(i < t)$ is available at task $t$, i.e., $\widetilde{N}_{i} \ll N_{i}$, and $\widetilde{N}_{i}$ is the replayed samples amount, leading the task $i$'s prediction error to deviate much from their overall real risks with severe forgetting issues.
\end{theorem}

However, in OWCL, when data from previous tasks is unavailable, leveraging the historical model $h^{t-1}$ from time $(t-1)$ instead of replaying a subset of data should be an alternative to tighten the prediction error bound, resulting in improved methods accordingly.
Hence, considering the most difficult OWCL scenario (i.e., KIRO), we derive \autoref{lemma1} and \autoref{lemma2} to unify a novel incremental prediction error bound for OWCL. Following the \autoref{the:task specific error}, we have:

\begin{lemma}\label{lemma1}
Let $h^t$ be the current function trained on task $t$ and $h^{t-1}$ is the model trained on the previous task $t-1$. Then, the $h^t$'s task-specific prediction error on an arbitrary task $i \leq t$ has an upper bound:
\begin{equation}\label{eq:task-specific bound1}
    \epsilon_{\mathcal{D}_{tr}^{i}}(h^t) \leq \epsilon_{\mathcal{D}_{tr}^{i}}\left(h^t, h^{t-1}\right)+\epsilon_{\mathcal{D}_{tr}^{i}}\left(h^{t-1}\right), 
\end{equation}
where $\epsilon_{\mathcal{D}_{tr}^{i}}\left(h^t, h^{t-1}\right) \triangleq \mathbb{E}_{\boldsymbol{x} \sim \mathcal{D}_{tr}^{i}}\left[h^t(\boldsymbol{x}) \neq h^{t-1}(\boldsymbol{x})\right]$. 
\end{lemma}
This lemma shows that the prediction error of the current model $h^t$’s on an arbitrary task $i$ is bounded by the difference between $h^t$ and its previous model $h^{t-1}$ plus the prediction error of $h^{t-1}$ on task $i$. 
Additionally, given the presence of $\epsilon_{\mathcal{D}_{tr}^{i}}\left(h^{t-1}\right)$, the task-specific prediction error bound of $h^{t-1}$ will be inherited to $\epsilon_{\mathcal{D}_{tr}^{i}}(h^t) $, which means the incremental prediction error will gradually accumulate as new task coming.
\begin{lemma}\label{lemma2}
Let $h^t$ be the current function trained on task $t$ and $h^{t-1}$ is the model trained on the previous task $t-1$. The task-specific prediction error on task $i (i \leq t)$ has an upper bound:
\begin{equation}
\begin{aligned}
        \epsilon_{\mathcal{D}_{tr}^t}(h^t) 
        & \leq \epsilon_{\mathcal{D}_{tr}^i}\left(h^t, h^{t-1}\right)\\
        & \ +\frac{1}{2} d_{\mathcal{H} \Delta \mathcal{H}}\left(\mathcal{D}_{tr}^i, \mathcal{D}_{tr}^t\right)+\epsilon_{\mathcal{D}_{tr}^i}\left(h^{t-1}\right),
\end{aligned}
\end{equation}
where $\mathcal{H}$ is a universal hypothesis space with finite VC dimension \cite{vapnik1998statistical}, and 
$d_{\mathcal{H} \Delta \mathcal{H}}(\mathcal{D}_{tr}^i, \mathcal{D}_{tr}^t)$ denotes the divergence between the distributions $\mathcal{D}_{tr}^i$  and  $\mathcal{D}_{tr}^t$. From a perspective of Bayesian theory, $d_{\mathcal{H} \Delta \mathcal{H}}(\mathcal{D}_{tr}^i, \mathcal{D}_{tr}^t)$ can be approximately calculated as
$2 \sup_{h^t \in \mathcal{H}} \left|\mathbf{P}_{\boldsymbol{x} \sim \mathcal{D}_{tr}^i}[h^t(\boldsymbol{x})=1]-\mathbf{P}_{\boldsymbol{x} \sim \mathcal{D}_{tr}^t}[h^t(\boldsymbol{x})=1]\right| $ to quantify the task-specific error. 
\end{lemma}

\autoref{lemma2} establishes a theoretical foundation for knowledge transfer by demonstrating that when the divergence between task \(i\) and the current task \(t\) (i.e., \(d_{\mathcal{H} \Delta \mathcal{H}}(\mathcal{D}_{tr}^i, \mathcal{D}_{tr}^t)\)) is sufficiently small, the predictions of \(h^{t-1}\) on \(\mathcal{D}_{tr}^t\) can effectively serve as a surrogate loss. This surrogate loss is crucial in mitigating forgetting by anchoring predictions on previously learned tasks. 
Based on this principle, many CL and OWCL methods design knowledge transfer mechanisms to minimize divergence, preserving task-specific knowledge while integrating new knowledge effectively.
Then, by using the task-specific prediction error bound in \autoref{lemma2}, we can derive the incremental prediction error bound by applying \autoref{lemma1} and \autoref{lemma2}:

\begin{theorem}\label{the:task incremental error bound}
\textbf{(Incremental Prediction Error Bound.) } With a probability of at least $1-\delta \text{\textgreater} 0$, the incremental prediction error bound can be refined as:
\begin{equation}\label{final task-incre bound}
    \begin{aligned}
\sum_{i=1}^{t} \epsilon_{\mathcal{D}_{tr}^{i}}(h^t) 
& \leq \left\{ \sum_{i=1}^{t-1} \left[\gamma_{i} \widehat{\epsilon}_{\mathcal{D}_{tr}^{i}}(h^t)
+ \alpha_{i} \widehat{\epsilon}_{\mathcal{D}_{tr}^{i}} \left(h^t, h^{t-1}\right) \right] \right\} \\
& \quad + \left\{ \widehat{\epsilon}_{\mathcal{D}_{tr}^t}(h^t) 
+ \left(\sum_{i=1}^{t-1} \beta_{i}\right) \widehat{\epsilon}_{\mathcal{D}_{tr}^t} \left(h^t, h^{t-1}\right) \right\} \\
& \quad + \frac{1}{2} \sum_{i=1}^{t-1} \beta_{i} d_{\mathcal{H} \Delta \mathcal{H}} \left( \mathcal{D}_{tr}^{i}, \mathcal{D}_{tr}^{t} \right) \\
& \quad + \sum_{i=1}^{t-1} \left(\alpha_{i} + \beta_{i}\right) \epsilon_{\mathcal{D}_{tr}^i} \left(h^{t-1}\right) + C_1 \\
& = \underbrace{\left\{ \sum_{i=1}^{t-1} \beta_{i} \widehat{\epsilon}_{\mathcal{D}_{tr}^t} \left(h^t, h^{t-1}\right) \right\}}_{\textbf{Model Discrepancy Terms}} \\
& \quad + \underbrace{\left\{ \sum_{i=1}^{t-1} \alpha_{i} \widehat{\epsilon}_{\mathcal{D}_{tr}^{i}} \left(h^t, h^{t-1}\right) \right\}}_{\textbf{Model Discrepancy Terms (cont.)}} \\
& \quad + \underbrace{\left\{ \sum_{i=1}^{t-1} \gamma_{i} \widehat{\epsilon}_{\mathcal{D}_{tr}^{i}}(h^t)
+ \widehat{\epsilon}_{\mathcal{D}_{tr}^t}(h^t) \right\}}_{\textbf{ERM-like Terms}} \\
& \quad + \underbrace{\left\{ \frac{1}{2} \sum_{i=1}^{t-1} \beta_{i} d_{\mathcal{H} \Delta \mathcal{H}} \left(\mathcal{D}_{tr}^{i}, \mathcal{D}_{tr}^{t}\right) \right\}}_{\textbf{Data Discrepancy Term}} + C_2.
    \end{aligned}
\end{equation}
where the constant $C_1$ can be calculated by ERM-Based generalization bound, $\gamma_{i}+\alpha_{i}+\beta_{i}=1$, and the error term $\sum_{i=1}^{t-1}\left(\alpha_{i}+\beta_{i}\right) \epsilon_{\mathcal{D}_{tr}^i}\left(h^{t-1}\right)$ is also a constant without any trainable parameters in the fixed (frozen) historical function $h^{t-1}$. The model discrepancy measures the weighted divergence between the current model $h^t$ and the previous one $h^{t-1}$, and the data discrepancy measures the scaled separability between the data distributions of task $i$ and task $t$.
\end{theorem}

\textbf{Proof. }
\begin{equation}
\begin{aligned}
\epsilon_{\mathcal{D}_{i}}(h) & = \left(\gamma_{i} + \alpha_{i} + \beta_{i}\right) \epsilon_{\mathcal{D}_{i}}(h) \\
& \leq \gamma_{i} \epsilon_{\mathcal{D}_{i}}(h) 
+ \alpha_{i} \left[\epsilon_{\mathcal{D}_{i}}\left(h, H_{t-1}\right) + \epsilon_{\mathcal{D}_{i}}\left(H_{t-1}\right)\right] \\
& \quad + \beta_{i} \left[\epsilon_{\mathcal{D}_{t}}\left(h, H_{t-1}\right) 
+ \frac{1}{2} d_{\mathcal{H} \Delta \mathcal{H}}\left(\mathcal{D}_{i}, \mathcal{D}_{t}\right) \right] \\
& \quad  + \beta_i \epsilon_{\mathcal{D}_{i}}\left(h^{t-1}\right),\\
\Rightarrow \sum_{i=1}^{t} \epsilon_{\mathcal{D}_{tr}^{i}}(h^t) 
& \leq \left\{\sum_{i=1}^{t-1}\left[\gamma_{i} \widehat{\epsilon}_{\mathcal{D}_{tr}^{i}}(h^t) 
+ \alpha_{i} \widehat{\epsilon}_{\mathcal{D}_{tr}^{i}}\left(h^t, h^{t-1}\right)\right]\right\} \\
& \quad + \left\{\widehat{\epsilon}_{\mathcal{D}_{t}}(h) 
+ \left(\sum_{i=1}^{t-1} \beta_{i}\right) \widehat{\epsilon}_{\mathcal{D}_{t}}\left(h, H_{t-1}\right)\right\} \\
& \quad + \frac{1}{2} \sum_{i=1}^{t-1} \beta_{i} d_{\mathcal{H} \Delta \mathcal{H}}\left(\mathcal{D}_{tr}^{i}, \mathcal{D}_{tr}^{t}\right) \\
& \quad + \sum_{i=1}^{t-1}\left(\alpha_{i} + \beta_{i}\right) \epsilon_{\mathcal{D}_{i}}\left(h^{t-1}\right) + C_1.
\end{aligned}
\end{equation}

Following \cite{anthony1999neural}, given each task $i$ has a deterministic ground-truth classifier $f^{i}$: $\mathbb{R}^{n} \rightarrow\{0,1\}$, the $C_1$ can be calculated by ERM-Based generalization bound:
\begin{equation}
\begin{aligned}
\sum_{i=1}^{t}  & \epsilon_{\mathcal{D}_{tr}^{i}}(h^t) 
 \leq \sum_{i=1}^{t} \widehat{\epsilon}_{\mathcal{D}_{tr}^{i}}(h^t) \\
& \quad + \sqrt{\left(\frac{\left(1 + \sum_{i=1}^{t-1} \beta_{i}\right)^{2}}{N_{t}} 
+ \sum_{i=1}^{t-1} \frac{\left(\gamma_{i} + \alpha_{i}\right)^{2}}{\widetilde{N}_{i}}\right)}\\
& \quad \times \sqrt{\left(8 d \log \left(\frac{2 e N}{d}\right) + 8 \log \left(\frac{2}{\delta}\right)\right)} \\
& \Rightarrow C_1 = \sqrt{\left(\frac{\left(1 + \sum_{i=1}^{t-1} \beta_{i}\right)^{2}}{N_{t}} 
+ \sum_{i=1}^{t-1} \frac{\left(\gamma_{i} + \alpha_{i}\right)^{2}}{\widetilde{N}_{i}}\right)} \\
& \quad \times \sqrt{\left(8 d \log \left(\frac{2 e N}{d}\right) + 8 \log \left(\frac{2}{\delta}\right)\right)}.
\end{aligned}
\end{equation}

Thus, $C_1$ is a constant with no trainable parameters and the \autoref{the:task incremental error bound} is proofed. $ \ \square$

Notably, our theory is compatible with all the incremental scenarios (TIL, DIL and CIL) and is capable of addressing all four OWCL settings we discussed in the introduction. In addition to the data discrepancy term, another crucial component in \autoref{the:task incremental error bound} is the model discrepancy term, which effectively characterizes the optimization direction and extensions (or isolations) of the previously trained model from $t-1$ to $t$. The model discrepancy term represents a fundamental aspect of current CIL design, particularly in regularization-based methods and dynamic structural models.
For the previous $t-1$ tasks, the bound is given by:
\begin{equation}
\begin{aligned}
    \epsilon_{\mathcal{D}_{tr}^i}(h^t) &\leq 
    \mathbb{E}_{h \sim Q^t} \left[ \widehat{\epsilon_{\mathcal{D}_{tr}^i}}(h) \right] \\
    &\quad + \sqrt{\frac{KL(Q^t \| P^{t-1}) + \log\frac{1}{\delta}}{2n_i}}, \\
    &\quad \text{for } i = 1, \dots, t-1.
\end{aligned}
\end{equation}
Summing these bounds across all tasks, we obtain:
\begin{equation}
\begin{aligned}
    \sum_{i=1}^{t} \epsilon_{\mathcal{D}_{tr}^i}(h^t) &\leq 
    \sum_{i=1}^{t} \mathbb{E}_{h \sim Q^t} \left[ \widehat{\epsilon_{\mathcal{D}_{tr}^t}}(h) \right] \\
    &\quad + \sum_{i=1}^{t} \sqrt{\frac{KL(Q^t \| P^{t-1}) + \log\frac{1}{\delta}}{2n_i}}.
\end{aligned}
\end{equation}
By decomposing, we get:
\begin{equation}
\begin{aligned}
    \sum_{i=1}^{t} \epsilon_{\mathcal{D}_{tr}^i}(h^t) &\leq 
    \sum_{i=1}^{t} \mathbb{E}_{h \sim Q^t} \left[ \widehat{\epsilon_{\mathcal{D}_{tr}^t}}(h) \right] \\
    &\quad + \sum_{i=1}^{t} \sqrt{\frac{KL(Q^t \| P^{t-1})}{2n_i}} \\
    &\quad + \sum_{i=1}^{t} \sqrt{\frac{\log\frac{1}{\delta}}{2n_i}}.
\end{aligned}
\end{equation}
Here, the model discrepancy $\sum_{i=1}^{t} \sqrt{\frac{KL(Q^t \| P^{t-1})}{2n_i}}$accounts for the divergence between the posterior distribution $Q^t$ and the prior $P^{t-1}$, normalized by the number of training samples $n_i$.

In this section, we theoretically refine an open risk in \autoref{the:open risk} and determine a tighter incremental prediction error bound in \autoref{the:task incremental error bound} that is more suitable for OWCL. 
Our theoretical analysis accounts for the knowledge of both knowns and unknowns, enhancing the separability of the knowledge space and facilitating the transfer of knowledge related to open samples. 

\section{Methodology}
\label{Method}
Motivated by empirical findings and theoretical analysis, the interaction between open risk and incremental prediction error drives us to approach the OWCL problem as an integrated whole.
Hence, we design a model that effectively transfers knowledge for both known categories and unknown samples, i.e., \textbf{knowns-unknowns knowledge transfer}.

From previous theoretical analysis, we have:

\begin{corollary}\label{coro2}
    Given a projection $\mathcal{P}: \mathbb{R} \rightarrow \mathbb{R}'$, an empirical estimate for the data discrepancy term is: 
\begin{equation}
\begin{aligned}
 & \frac{1}{2} \sum_{i=1}^{t-1} \beta_{i} d_{\mathcal{H} \Delta \mathcal{H}}\left(\mathcal{D}_{tr}^{i}, \mathcal{D}_{tr}^{t}\right) \\
 & = \frac{1}{2} \sum_{i=1}^{t-1} \beta_{i} \widehat{d}_{\mathcal{H} \Delta \mathcal{H}}\left(\mathcal{P} \left(\mathcal{B}^{i}\right), \mathcal{P}\left(\mathcal{B}^{t}\right)\right) \\
 & = \sum_{i=1}^{t-1} \beta_{i} \left( 1- \min _{d \in \mathcal{H}_{d}}\left[\frac{1}{\widetilde{N}_{i}} \sum_{\boldsymbol{x} \in \mathcal{X}_{i}} \left[-\log \left([d(\mathcal{P}(\boldsymbol{x}))]_{i}\right)\right] \right. \right. \\
 & \left. \left. \qquad + \frac{1}{N_{t}} \sum_{\boldsymbol{x} \in \mathcal{S}_{t}} \left[-\log \left([d(\mathcal{P}(\boldsymbol{x}))]_{t}\right)\right] \right] \right),
\end{aligned}
\end{equation}
where $\mathcal{B}^{t}$ and $\mathcal{B}^{i}$ are experience buffers drawn from training set $\mathcal{D}_{tr}^{t}$ and $\mathcal{D}_{tr}^{i}$ of tasks $t$ and $i$, respectively. 
\end{corollary}

Given the most difficult scenario, KIRO, each task’s training set comes incrementally with repeated classes and changing distributions, which may result in overlapping across different tasks' semantic spaces, thus increasing the \textbf{\textit{inter-task open risk}}.
Additionally, the repeated appearance of open samples may increase the \textbf{\textit{intra-task open risk}} by identifying open/unknown samples as known.
Therefore, it is not enough to pull together all seen/known samples belonging to the same class, but we need to \textit{(1) find a proper projection with improved linear separability} and \textit{(2) determine appropriate experience buffers to describe embedding distributions of different tasks.}

Subsequently, we propose a novel OWCL framework that can jointly learn knowledge about known classes and unknown samples, effectively generalizing under open-world assumptions while excelling in both open detection and incremental classification, termed HoliTrans.
HoliTrans can infer general patterns from specific observed instances, which is crucial for adapting to new classes and detecting novel samples within all the OWCL scenarios.

\subsection{Initialization: Fine-tuning on the First Task}
Given a sequence of tasks, the training data and labels from the first task may better represent the subsequent tasks than the data used to train the original pre-trained model. 
Hence, we first need to fine-tune the pre-trained model via a Parameter-Efficient Transfer Learning (PETL) strategy \cite{houlsby2019parameter,li2021prefix}.
Moreover, due to the advantages of the random projection (RP) layer being independent of the model, it can be applied to any feature extractor.
Therefore, it can be applied orthogonally to the widely applied PETL strategy, which does not alter any parameters of the original pre-trained model. 

In this work, we fine-tune the pre-trained model with the first task by learning PETL parameters and then freeze them for the subsequent tasks.
Here, we conduct three competitive PETL methods: AdaptFormer \cite{chen2022adaptformer}, SSF \cite{lian2022scaling}, and VPT \cite{jia2022visual}.
After fine-tuning the pre-trained model with the first task, we conduct a novel two-stage training approach for each task (as shown in \autoref{algo1}).

\subsection{The Proposed Method}
Given a new task $t$, we extract features from its training set using the fine-tuned pre-trained model and obtain the corresponding embeddings.
In the first stage training, we introduce the \textbf{nonlinear random projection (NRP)} from the assumption that nonlinear random interactions between embedding distributions may be more linearly separable than the original in a higher-level embedding space, with corresponding mathematical properties that have been discussed extensively \cite{chen1996rapid,huang2006extreme,mcdonnell2024ranpac}.
Thus, the embeddings can be improved as:
\begin{equation}
    H^i = g(\phi(D^i_{tr})W).
\end{equation}

Then, we encode the labels into a one-hot format, obtaining the label matrix $Y^i$. Accordingly, we update the model's Gram matrix and class prototype matrix as follows:

\begin{equation}\label{updateGandC}
\begin{aligned}
        G^i &= G^{i - 1} + H^{i^T}H^i, \\
    C^i &= C^{i - 1} + H^{i^T}Y.
\end{aligned}
\end{equation}

Given the training set of task $t$, each class's prototype can be acquired from \autoref{updateGandC}. 
Since the prototype for each known category is updated through NPR, we refer to these prototypes as \textbf{Distribution-Aware Prototypes (DAPs)}.
Additionally, we compute the mean-variance of distances between samples of all classes and their prototypes:

\begin{equation}
    \label{updatedelta}
    \delta^i = \frac{\delta^{i-1}\sum_{j = 1}^{i - 1}N_j + \mathbb{E}[(H^i - H^{i^T})^2]}{\sum_{j = 1}^iN_j}.
\end{equation}

To enable HoliTrans to acquire knowledge from known classes without forgetting prior tasks, we store the learned DAPs incrementally. 
Given that OWCL presents greater challenges, it is vital not only to reduce incremental classification error but also to establish tighter decision boundaries for open detection in testing. 
Therefore, we propose a novel pseudo-sample generation method and utilize these pseudo-samples to enhance the test set (in contrast to all replay-based CL methods, which incorporate generated samples into the training set), to make full use of the knowledge of knowns and opens.

Given a specific known class $k$ with its DAP $\boldsymbol{p}_k$, we generate \textit{positive pseudo-samples} that follow the distribution $\sim(\boldsymbol{p}_{k}, \delta^2)$ to ensure that the positive pseudo-samples are centered around $\boldsymbol{p}_k$.
Moreover, we generate random \textit{negative pseudo-samples} that follow a distribution $\sim(\boldsymbol{p}_{kl}, \delta^2)$ at arbitrary positions between each pair of prototypes $(\boldsymbol{p}_k, \boldsymbol{p}_l)$, where $\boldsymbol{p}_{kl} = \frac{\boldsymbol{p}_k + \zeta \boldsymbol{p}_l}{1 + \zeta}$.
These random \textit{negative pseudo-samples} form the pseudo-test set $\mathcal{D'}_{te}^i$.

\subsection{Overall Optimization}
Subsequently, we propose a novel threshold learning approach with knowledge-adaptive capability, building on the NCM-based classifier. Our goal is to determine an appropriate threshold for each task, enabling the classifier to discern whether a test sample belongs to a known category. If it does, the NCM-based classifier aligns the sample with the corresponding DAP's label and gets the output. If not, the test sample is identified as open.

Consequently, we design the following optimization problem to determine the threshold \( r \):

\begin{align}\label{eq:r}
    \arg & \max_{r} \frac{1}{|\mathcal{D'}_{te}^i|} \sum_{\boldsymbol{x} \in \mathcal{D'}_{te}^i}  
    \left[ \mathbb I\left(\boldsymbol{s_x} < r \frac{1}{|\mathcal{D'}_{tr}^i|} \sum_{\boldsymbol{z} \in \mathcal{D'}_{tr}^i} \boldsymbol{s_z} \, | \, \text{known}\right) \right. \nonumber \\
    & \quad + \left. \mathbb I\left(\boldsymbol{s_x} > r \frac{1}{|\mathcal{D'}_{tr}^i|} \sum_{\boldsymbol{z} \in \mathcal{D'}_{tr}^i} \boldsymbol{s_z} \, | \, \text{open}\right) \right] \\
    & \text{\ \ with\ \ } \boldsymbol{s_x} = \max_{y \in \mathcal{Y}^{M_i}} g\left(\phi(x)^T W\right) (G^i + \lambda^i I)^{-1} \boldsymbol{p}_y^i, \nonumber \\
    & \quad \text{and\ \ } \boldsymbol{p}_y^i = \frac{\sum_{j = 1}^{|\mathcal{X}^i|} \mathbb I(y^i_j = y) \phi(x^i_j)}{\sum_{j = 1}^{|\mathcal{X}^t|} \mathbb I(y^i_j = y)}.
\end{align}
where $\mathbb{I}(\cdot)$ is an indicator function, $\boldsymbol{s_x} $ is the maximum similarity between sample $x$ with learned prototypes, $g(\cdot)$ is a nonlinear function, $\phi(\cdot)$ is the pre-trained model, $G^i$ is the Gram matrix of NRP feature, and $\boldsymbol{p}^i_y$ is the prototype of class $y$ in task $i$. Note the $\mathcal{D'}_{te}^i$ is different from the original $\mathcal{D}_{te}^i$ in testing: we integrate a batch of generated pseudo-samples for each task to better represent data distribution via the embedding distribution over $\mathcal{D}_{tr}^i$, obtaining DAP for each known class.

Naturally, in solving for \( r \), we need further analysis as:
\begin{equation}
\label{function}
\begin{split}
    \frac{1}{|\mathcal{D'}_{te}^i|} \sum_{\boldsymbol{x} \in \mathcal{D'}_{te}^i} &\left[ \mathbb{I}\left(\boldsymbol{s_x} < r \frac{1}{|\mathcal{D'}_{tr}^i|} \sum_{\boldsymbol{z} \in \mathcal{D'}_{tr}^i} \boldsymbol{s_z} \mid \text{known}\right) \right. \\
    &\quad + \left. \mathbb{I}\left(\boldsymbol{s_x} > r \frac{1}{|\mathcal{D'}_{tr}^i|} \sum_{\boldsymbol{z} \in \mathcal{D'}_{tr}^i} \boldsymbol{s_z} \mid \text{open}\right) \right],
\end{split}
\end{equation}
exhibits the properties of being \textbf{\textit{unimodal, long-tailed, and decreasing}}. Consequently, in our HoliTrans framework, we employ the classic numerical technique of ternary search, which adaptively identifies the optimal threshold $r$. The ternary search algorithm iteratively refines the search interval to determine the parameter value that maximizes accuracy for the NCM-based classifier.

The \autoref{algo1} summarizes the overall training process.
The overall training process consists of two main phases. In the first phase, given a task \(i\) with its training set \(D_{tr}^i\) and test set \(D_{te}^i\), the class prototypes \(G^i\) and \(\{\boldsymbol{p_k}\}_{k=1}^{M^i}\) are updated using the training data, and the standard deviation \(\delta\) between samples and their corresponding prototypes is calculated. In the second phase, the algorithm generates discriminative auxiliary pseudo-samples (DAPs). Negative pseudo-samples are created by computing pseudo-prototypes from pairs of prototypes \((\boldsymbol{p}_k, \boldsymbol{p}_l)\) and sampling from a Gaussian distribution centered around the pseudo-prototype. Positive pseudo-samples are similarly generated from each prototype \(\boldsymbol{p_k}\) using a Gaussian distribution. The pseudo-samples are then integrated with the test set \(\mathcal{D}_{te}^i\), and the decision threshold \(r\) is determined through a ternary search over the modified test set \(\mathcal{D'}_{te}^i\). This process enhances the classifier's ability to distinguish between known and open classes by utilizing both positive and negative pseudo-samples.
      \begin{algorithm}[H]
        \caption{Training Process} \label{algo1}
            \SetKwInOut{Input}{Input}
            \SetKwInOut{Output}{Output}
            \Input{Given a task $i$ with its training set $D_{tr}^i$ and test set $D_{te}^i$.}
                \textbf{Phase 1: } \textit{NCM-based Classifier Training}\\
                Update $G^i$ and $\{\boldsymbol{p_k}\}^{k=M^i}_{k=1}$ by $D_{tr}^i$;\\
                Calculate a standard deviation $\delta$ between samples and corresponding prototypes.\\
                    
            \textbf{Phase 2: } \textit{Learning DAPs}

                \tcp{Generate Negative Pseudo-samples}
                
                 \For{an arbitrary pair of prototypes $(\boldsymbol{p}_k,\boldsymbol{p}_l)$}{
                    Obtain seudo-prototype $\boldsymbol{p}_{kl} = \frac{\boldsymbol{p}_k + \zeta \boldsymbol{p}_l}{1 + \zeta}$;\\
                    Generate negative pseudo-samples via a Gaussian distribution $\sim(\boldsymbol{p}_{kl} , {\delta^i}^2)$.
                }

                \tcp{Generate Positive Pseudo-samples}

                \For{each prototype $\boldsymbol{p_k}$}{
                    Generate positive pseudo-samples via a Gaussian distribution $\sim(\boldsymbol{p}_{k} , {\delta^i}^2)$.
                }
                Integrating all pseudo-samples with $\mathcal{D}_{te}^i$;\\
                Determining $r$ by ternary search over $\mathcal{D'}_{te}^i$.
                
      \end{algorithm}

\subsection{Complexity Analysis} 
The complexity of generating pseudo-samples is $O(n^2\times dim)$, the time consuming of evaluating pseudo-samples is $O(dim\times M\times n)$, where $n$ is the number of classes, $dim$ is the dimension of the pre-trained model output and $M$ is the NRP size.
The complexity of solving the optimal solution for $r$ in DAP is $O(log_3(\frac{\max_{\boldsymbol{ z} \in \mathcal{D}_{tr}^i}\boldsymbol{ s_z} - \min_{\boldsymbol{z} \in \mathcal{D}_{tr}^i}\boldsymbol{ s_z}}{\frac{1}{|\mathcal{D}_{tr}^i|} \sum_{\boldsymbol{ z} \in \mathcal{D}_{tr}^i} \boldsymbol{ s_z}}  / \epsilon))$, where $\epsilon$ is an error rate for terminating the ternary earching.

\subsection{Why NPR performs effectively in Knowledge Transfer for OWCL}
To understand why NRP performs effectively in OWCL, we provide extra theoretical analyses as follows.

First, NRP approximates each class distribution to a Gaussian distribution. 
Given a vector \(f\), the norm of its projected vector can be bounded using the Chernoff bound:

\begin{equation}
\begin{aligned}
    \mathbb{P} \big(|\|W^T f\| - \mathbb{E}_W[\|W^T f\|]| 
    &> \epsilon\sigma^2 \big) \\
    &\leq 2e^{-\frac{\epsilon^2\sigma^2}{2M + \epsilon}}.
\end{aligned}
\end{equation}
This bound reveals the relationship between dimensionality and the expected variation in the norm of the projected vector. For fixed \(\epsilon\) and \(\sigma\), as \(M\) increases, the probability on the right-hand side approaches 1. This indicates that the norm of the projected vector is more likely to fall within an ideal range around the expected value. In other words, in higher dimensions, these projected vectors tend to lie near the boundaries of the distribution, with similar distances from the mean—making the distribution align more closely with a Gaussian.

Secondly, NRP can lead to better decision boundaries. Consider any two vectors \(f\) and \(f'\), and analyze the Chernoff bound for their inner product before and after projection:

\begin{equation}
\begin{aligned}
    \mathbb{P} \big( |(W^T f)^T (W^T f') 
    &- \mathbb{E}_W[(W^T f)^T (W^T f')]| \\
    &> \epsilon'M\sigma^2 \big) \leq 2e^{-\frac{\epsilon'^2\sigma^2}{2M + \epsilon'}}.
\end{aligned}
\end{equation}
This can be rewritten as:
\begin{equation}
\begin{aligned}
    \mathbb{P} \big( |(W^T f)^T (W^T f') 
    &- M\sigma^2 f^T f'| \\
    &> \epsilon'M\sigma^2 \big) \leq 2e^{-\frac{\epsilon'^2\sigma^2}{2M + \epsilon'}}.
\end{aligned}
\end{equation}
Simplifying further:
\begin{equation}
\begin{aligned}
    \mathbb{P} \bigg( \bigg| \frac{(W^T f)^T (W^T f')}{M\sigma^2} 
    &- f^T f' \bigg| > \epsilon' \bigg) \\
    &\leq 2e^{-\frac{\epsilon'^2\sigma^2}{2M + \epsilon'}}.
\end{aligned}
\end{equation}

This bound shows that as the projection dimensionality \(M\) increases, the inner products of arbitrary vectors and their projections are less likely to be equal. In other words, higher dimensions reduce the probability that two vectors and their projections share the same inner product. This characteristic facilitates the establishment of better decision boundaries for OWCL.

\section{Experiments}
\label{Experiments and Results}

\subsection{Experimental Settings}
\subsubsection{Baselines} 
We compare HoliTrans with \textbf{22} baseline models, including OWCL methods and various continual learning (CL) methods with several out-of-distribution (OOD) algorithms.
We consider \textbf{5} state-of-the-art CL methods that leverage large-scaled pre-training and prompt-tuning as well. 
Below is a brief summary of each baseline model:

\begin{itemize}
    \item \textbf{L2P} \cite{wang2022learning}: This method uses prompt-based fine-tuning to adapt pre-trained models to continual learning tasks efficiently, leveraging large-scale pre-trained knowledge.
    
    \item \textbf{DualPrompt} \cite{wang2022dualprompt}: This approach introduces dual prompt tuning, focusing on both task-specific and task-agnostic prompts to enhance generalization in continual learning scenarios.
    
    \item \textbf{CODA} \cite{smith2023coda}: CODA leverages contrastive learning and dynamic prompts to improve performance across continual learning tasks by preserving learned knowledge while adapting to new tasks.
    
    \item \textbf{RanPAC} \cite{mcdonnell2024ranpac}: A memory-efficient continual learning method that uses randomized partial attention to capture relevant task-specific knowledge while mitigating catastrophic forgetting.
    
    \item \textbf{ADAM} \cite{zhou2023revisiting}: ADAM revisits the attention-based learning paradigm in continual learning, proposing an architecture that better preserves previous task knowledge through dynamic attention mechanisms.
    
    \item \textbf{OpenMax} \cite{bendale2015towards}: This OOD detection method modifies the final layer of deep networks to identify unknown classes by adjusting softmax outputs to better handle open-set recognition.
    
    \item \textbf{MaxLogits} \cite{basart2022scaling}: MaxLogits utilizes the maximum logit score to identify out-of-distribution samples, offering a simple and scalable OOD detection technique for continual learning.
    
    \item \textbf{Entropy} \cite{chan2021entropy}: This method employs the entropy of the output distribution as a metric to distinguish between in-distribution and out-of-distribution data for OOD detection.
    
    \item \textbf{EnergyBased} \cite{liu2020energy}: Energy-based OOD detection uses the energy score derived from the output of the neural network to separate in-distribution samples from OOD instances, improving recognition of unseen classes.
    
    \item \textbf{MORE} \cite{kim2022theoretical}: MORE is an OWCL method that introduces memory-efficient regularization techniques to better handle task transitions while preventing forgetting.
    
    \item \textbf{Pro-KT} \cite{li2024learning}: Pro-KT is a prototype-based continual learning method that facilitates knowledge transfer across tasks, utilizing prototypes to manage open-world scenarios effectively.
\end{itemize}

\subsubsection{Datasets and Scenario Setups} As mentioned earlier, CIRO and KIRO are the two more challenging OWCL scenarios.
For the CIRO scenario, we divide the \textbf{Split-CIFAR100} dataset \cite{krizhevsky2009learning} into 10 tasks (10 classes for each task) and conduct testing using the entire testing set of all 10 tasks. For the KIRO scenario, we use the object recognition dataset, \textbf{Open-CORe50} \cite{lomonaco2017core50}. In this dataset, new training samples of the same classes become available in a sequence of tasks with changing distributions (e.g., new conditions and backgrounds). Additionally, we specifically designate a certain class as open during testing.

As shown in \autoref{Dataset}, in the CIFAR100 dataset in the CIRO scenario, there are 90 known classes ($M_{known}$) and 10 unknown classes ($M_{open}$), with a total of 9 tasks and 55,000 samples. Each training epoch consists of 5000 samples ($N_t$), where $t = 1, \ldots, T$. The validation set comprises 9000 samples from known classes ($N_{val, known}$) and 1000 samples from unknown classes ($N_{val, open}$).

In the CORe50 dataset in the KIRO scenario, there are 25 known classes and 25 unknown classes, with 4 tasks and a total of 59,936 samples. The number of samples in each training epoch is as follows: $N_1 = 14,989$, $N_2 = 14,986$, $N_3 = 14,995$, and $N_4 = 14,966$. The validation set comprises 7528 samples from known classes and 7466 samples from unknown classes.

\begin{table*}[ht]
\centering
\caption{Dataset Descriptions.}
\label{Dataset}
\begin{tabular}{cccccccc}
\toprule
Dataset  & $M_{known}$ & $M_{open}$ & $T$ & $N$  & $N_t,t = 1,...,T$ & $N_{val,known}$ & $N_{val,open}$ \\ \toprule
CIFAR100 & \cellcolor{gray!20}90          & 10         & \cellcolor{gray!20}9   & 55000 & \cellcolor{gray!20}$N_t = 5000,t = 1,...,T$ & 9000            & \cellcolor{gray!20}1000           \\ \hline
CORe50   & \cellcolor{gray!20}25          & 25         & \cellcolor{gray!20}4   & 59936 & \cellcolor{gray!20}\begin{tabular}[c]{@{}c@{}}$N_1 = 14989$\\ $N_2 = 14986$\\ $N_3 = 14995$\\ $N_4 = 14966$\end{tabular} & 7528            & \cellcolor{gray!20}7466           \\ \bottomrule
\end{tabular}
\end{table*}

\subsubsection{Implementations}
The proposed HoliTrans is implemented based on PyTorch and released in supplemental materials. All experiments are conducted on a single NVIDIA RTX 3090 GPU. The utilized pre-trained backbone is ViT-B/16, which is self-supervised and pre-trained on ImageNet-21K. In the testing phase of each task, we treat the test samples of classes that have not yet been learned as open/unknown classes to conduct the CIRO and KIRO scenarios, i.e., the two more challenging and difficult scenarios of OWCL.

For the first stage of HoliTrans, we employ SGD to train the parameters of the PETL method, namely AdaptFormer, SSF, and VPT. For each of these, we use a batch size of 48, a learning rate of 0.01, weight decay of 0.0005, momentum of 0.9, and cosine annealing with restarts ending at a learning rate of 0. Throughout, we typically train for 20 epochs, although in some experiments, this may be reduced if overfitting becomes apparent. Softmax and cross-entropy loss are utilized when employing these methods. The number of classes equals the quantity from the first task, i.e., \( M_1 \). Before commencing the second stage of HoliTrans, we discard the training weights and heads generated thereby.

During training, data augmentation for all datasets includes random resizing followed by cropping to 224x224 pixels and random horizontal flipping. For inference, images are resized to a short side of 256 pixels and then center-cropped to 224x224 for all datasets except CIFAR100, which is directly resized from its original 32x32 size to 224x224.
\begin{table*}[htb]
\renewcommand\arraystretch{1}
\centering
\caption{ Main Results. We format \textbf{first}, \underline{second} and \underline{third} performances.}\label{tab:main results}
\vspace{2mm}
\begin{tabular}{cccccccc}
\toprule
\multicolumn{2}{c}{\multirow{2}{*}{\textit{\textbf{Method}}}} & \multicolumn{3}{c}{\textbf{Scenario: CIRO (Split-CIFAR100)}}                                 & \multicolumn{3}{c}{\textbf{Scenario: KIRO (Open-CORe50)}}               \\ \cmidrule(lr){3-5}\cmidrule(lr){6-8}
\multicolumn{2}{c}{}                                 & $ACC_t(\uparrow)$                        & $AUC_t (\uparrow)$ & $FPR_t (\downarrow)$ & $ACC_t (\uparrow)$   & $AUC_t (\uparrow)$   & $FPR_t (\downarrow)$ \\ \hline
 \multicolumn{8}{c}{\textit{Continual learning methods equipped with OOD algorithms}} \\ \hline
    \multirow{4}{*}{L2P}              & OpenMax          & $0.436_{\pm 0.048}$                          & $0.133_{\pm 0.051}$    & $0.937_{\pm 0.046}$      & $0.424_{\pm 0.063}$      & $0.244_{\pm 0.050}$      & $0.968_{\pm 0.039}$      \\
                                      & MaxLogits        & $0.445_{\pm 0.041}$                          & $0.103_{\pm 0.047}$    & $0.931_{\pm 0.045}$      & $0.412_{\pm 0.179}$      & $0.259_{\pm 0.053}$      & $0.964_{\pm 0.037}$      \\
                                      & Entropy          & $0.445_{\pm 0.028}$                          & $0.120_{\pm 0.038}$    & $0.932_{\pm 0.041}$      & $0.424_{\pm 0.063}$      & $0.257_{\pm 0.063}$      & $0.958_{\pm 0.046}$      \\
                                      & EnergyBased      & $0.445_{\pm 0.038}$                          & $0.102_{\pm 0.043}$    & $0.925_{\pm 0.040}$      & $0.425_{\pm 0.065}$      & $0.256_{\pm 0.057}$      & $0.970_{\pm 0.057}$      \\ \hline
    \multirow{4}{*}{DualPrompt}       & OpenMax          & $0.423_{\pm 0.031}$                          & $0.139_{\pm 0.037}$    & $0.928_{\pm 0.045}$      & $0.402_{\pm 0.029}$      & $0.270_{\pm 0.024}$      & $0.960_{\pm 0.036}$      \\
                                      & MaxLogits        & $0.425_{\pm 0.034}$                          & $0.114_{\pm 0.039}$    & $0.933_{\pm 0.042}$      & $0.408_{\pm 0.113}$      & $0.261_{\pm 0.017}$      & $0.975_{\pm 0.022}$      \\
                                      & Entropy          & $0.425_{\pm 0.045}$                          & $0.204_{\pm 0.032}$    & $0.936_{\pm 0.048}$      & $0.408_{\pm 0.112}$      & $0.267_{\pm 0.026}$      & $0.961_{\pm 0.019}$      \\
                                      & EnergyBased      & $0.425_{\pm 0.046}$                          & $0.114_{\pm 0.049}$    & $0.942_{\pm 0.039}$      & $0.408_{\pm 0.112}$      & $0.260_{\pm 0.030}$      & $0.973_{\pm 0.020}$      \\ \hline
    \multirow{4}{*}{CODA}             & OpenMax          & $0.439_{\pm 0.036}$                          & $0.125_{\pm 0.042}$    & $0.936_{\pm 0.046}$      & $0.413_{\pm 0.032}$      & $0.278_{\pm 0.045}$      & $0.953_{\pm 0.020}$      \\
                                      & MaxLogits        & $0.439_{\pm 0.047}$                          & $0.090_{\pm 0.039}$    & $0.942_{\pm 0.045}$      & $0.415_{\pm 0.033}$      & $0.183_{\pm 0.028}$      & $0.964_{\pm 0.018}$      \\
                                      & Entropy          & $0.439_{\pm 0.036}$                          & $0.106_{\pm 0.046}$    & $0.939_{\pm 0.047}$      & $0.415_{\pm 0.033}$      & $0.256_{\pm 0.056}$      & $0.953_{\pm 0.021}$      \\
                                      & EnergyBased      & $0.439_{\pm 0.031}$                          & $0.089_{\pm 0.053}$    & $0.946_{\pm 0.037}$      & $0.415_{\pm 0.033}$      & $0.169_{\pm 0.022}$      & $0.962_{\pm 0.018}$      \\ \hline
    \multirow{4}{*}{ADAM}             & OpenMax          & $0.440_{\pm 0.044}$                          & $0.112_{\pm 0.040}$    & $0.876_{\pm 0.043}$      & $0.433_{\pm 0.025}$      & $0.283_{\pm 0.038}$      & $0.960_{\pm 0.029}$      \\
                                      & MaxLogits        & $0.440_{\pm 0.042}$                          & $0.099_{\pm 0.053}$    & $0.882_{\pm 0.033}$      & $0.433_{\pm 0.026}$      & $0.238_{\pm 0.029}$      & $0.965_{\pm 0.025}$      \\
                                      & Entropy          & $0.440_{\pm 0.042}$                          & $0.221_{\pm 0.043}$    & $0.876_{\pm 0.047}$      & $0.433_{\pm 0.025}$      & $0.197_{\pm 0.016}$      & $0.967_{\pm 0.021}$      \\
                                      & EnergyBased      & $0.440_{\pm 0.035}$                          & $0.386_{\pm 0.042}$    & $0.869_{\pm 0.041}$      & $0.433_{\pm 0.027}$      & $0.345_{\pm 0.023}$      & $0.906_{\pm 0.035}$      \\ \hline
    \multirow{4}{*}{RanPAC}           & OpenMax          & $0.467_{\pm 0.045}$                          & $0.091_{\pm 0.044}$    & $0.879_{\pm 0.054}$      & $0.472_{\pm 0.022}$      & $0.201_{\pm 0.020}$      & $0.952_{\pm 0.013}$      \\
                                      & MaxLogits        & $0.468_{\pm 0.042}$                          & $0.092_{\pm 0.042}$    & $0.878_{\pm 0.039}$      & $0.472_{\pm 0.022}$      & $0.212_{\pm 0.017}$      & $0.963_{\pm 0.015}$      \\
                                      & Entropy          & $0.468_{\pm 0.049}$ & $0.091_{\pm 0.041}$    & $0.871_{\pm 0.031}$      & $0.472_{\pm 0.022}$      & $0.219_{\pm 0.021}$      & $0.962_{\pm 0.010}$      \\
                                      & EnergyBased      & $0.468_{\pm 0.046}$                          & $0.227_{\pm 0.038}$    & $0.869_{\pm 0.041}$      & $0.475_{\pm 0.021}$      & $0.411_{\pm 0.025}$      & $0.939_{\pm 0.021}$      \\ \hline
    \multicolumn{8}{c}{\textit{Open-world continual learning methods}} \\ \hline
    \multicolumn{2}{c}{MORE}                                    & \multicolumn{1}{c}{\underline{$0.716_{\pm 0.011}$}} & \multicolumn{1}{c}{\underline{$0.717_{\pm 0.127}$}} & \underline{$0.492_{\pm 0.011}$} & \multicolumn{1}{c}{\underline{$0.641_{\pm 0.077}$}}                & \multicolumn{1}{c}{\underline{$0.641_{\pm 0.077}$}}                & \multicolumn{1}{c}{\underline{$0.517_{\pm 0.011}$}} \\ 

    \multicolumn{2}{c}{Pro-KT}                                     & \multicolumn{1}{c}{\underline{$0.779_{\pm 0.010}$}} & \multicolumn{1}{c}{\underline{$0.745_{\pm 0.011}$}} & \underline{$0.397_{\pm 0.026}$} & \multicolumn{1}{c}{\underline{$0.635_{\pm 0.011}$}}                & \multicolumn{1}{c}{\underline{$0.675_{\pm 0.125}$}}                & \multicolumn{1}{c}{\underline{$0.451_{\pm 0.015}$} }\\
    \multicolumn{2}{c}{\textbf{HoliTrans}}                                     & \multicolumn{1}{c}{$\boldsymbol{0.851_{\pm0.010}}$} & \multicolumn{1}{c}{$\boldsymbol{0.858_{\pm0.012}}$} & $\boldsymbol{0.193_{\pm0.024}}$ & \multicolumn{1}{c}{$\boldsymbol{0.748_{\pm0.010}}$}                & \multicolumn{1}{c}{$\boldsymbol{0.748_{\pm0.010}}$}                & \multicolumn{1}{c}{$\boldsymbol{0.097_{\pm0.031}}$} \\ 
\bottomrule
\end{tabular}
\end{table*}

\begin{table*}[htb]
\renewcommand\arraystretch{1}
\centering
\caption{ Main Results. We format \textbf{first}, \underline{second} and \underline{third} performances.}
\vspace{2mm}
\begin{tabular}{cccccccc}
\toprule
\multicolumn{2}{c}{\multirow{2}{*}{\textit{\textbf{Method}}}} & \multicolumn{3}{c}{\textbf{Scenario: CIRO (Split-CIFAR100)}}                                 & \multicolumn{3}{c}{\textbf{Scenario: KIRO (Open-CORe50)}}               \\ \cmidrule(lr){3-5}\cmidrule(lr){6-8}
\multicolumn{2}{c}{}                                 & $ACC_t(\uparrow)$                        & $AUC_t (\uparrow)$ & $FPR_t (\downarrow)$ & $ACC_t (\uparrow)$   & $AUC_t (\uparrow)$   & $FPR_t (\downarrow)$ \\ \hline
    \multirow{4}{*}{RanPAC + LOF}              & 100\%          & $0.582_{\pm 0.009}$                          & $0.575_{\pm 0.003}$    & $0.795_{\pm 0.007}$      & $0.505_{\pm 0.005}$      & $0.502_{\pm 0.005}$      & $0.459_{\pm 0.004}$      \\ 
                                      & 75\%        & $0.584_{\pm 0.004}$                          & $0.565_{\pm 0.005}$    & $0.778_{\pm 0.002}$      & $0.501_{\pm 0.001}$      & $0.502_{\pm 0.004}$      & $0.466_{\pm 0.013}$      \\
                                      & 50\%          & $0.589_{\pm 0.009}$                          & $0.564_{\pm 0.004}$    & $0.699_{\pm 0.045}$      & $0.504_{\pm 0.002}$      & $0.501_{\pm 0.002}$      & $0.478_{\pm 0.004}$      \\
                                      & 25\%      & $0.583_{\pm 0.015}$                          & $0.566_{\pm 0.014}$    & $0.642_{\pm 0.049}$      & $0.509_{\pm 0.002}$      & $0.509_{\pm 0.003}$      & $0.467_{\pm 0.007}$      \\ \hline
                                      \multicolumn{2}{c}{\textbf{HoliTrans}}                                     & \multicolumn{1}{c}{$\boldsymbol{0.851_{\pm0.010}}$} & \multicolumn{1}{c}{$\boldsymbol{0.858_{\pm0.012}}$} & $\boldsymbol{0.193_{\pm0.024}}$ & \multicolumn{1}{c}{$\boldsymbol{0.748_{\pm0.010}}$}                & \multicolumn{1}{c}{$\boldsymbol{0.748_{\pm0.010}}$}                & \multicolumn{1}{c}{$\boldsymbol{0.097_{\pm0.031}}$}\\
\bottomrule
\end{tabular}
\end{table*}

\subsubsection{Metrics} 
We employ three key metrics, namely $ACC_t$, $AUC_t$, and $FPR_t$, to provide a comprehensive evaluation of model performance across $T$ tasks. 
Specifically, $ACC_t$ represents the average final accuracy concerning all previously encountered classes across the $t$ tasks. It evaluates how well the model retains knowledge from earlier tasks without significant forgetting. $AUC_t$ denotes the average area under the receiver operating characteristic (ROC) curve over all past tasks, reflecting the model’s ability to distinguish between known and unknown instances, thereby offering insights into its classification reliability. $FPR_t$, or the average false positive rate, quantifies the error rate in open-set detection, indicating how often the model mistakenly classifies unknown instances as belonging to known categories.

\subsection{Results Analysis}

\begin{figure*}[ht]\centering
\includegraphics[width=0.8\linewidth]{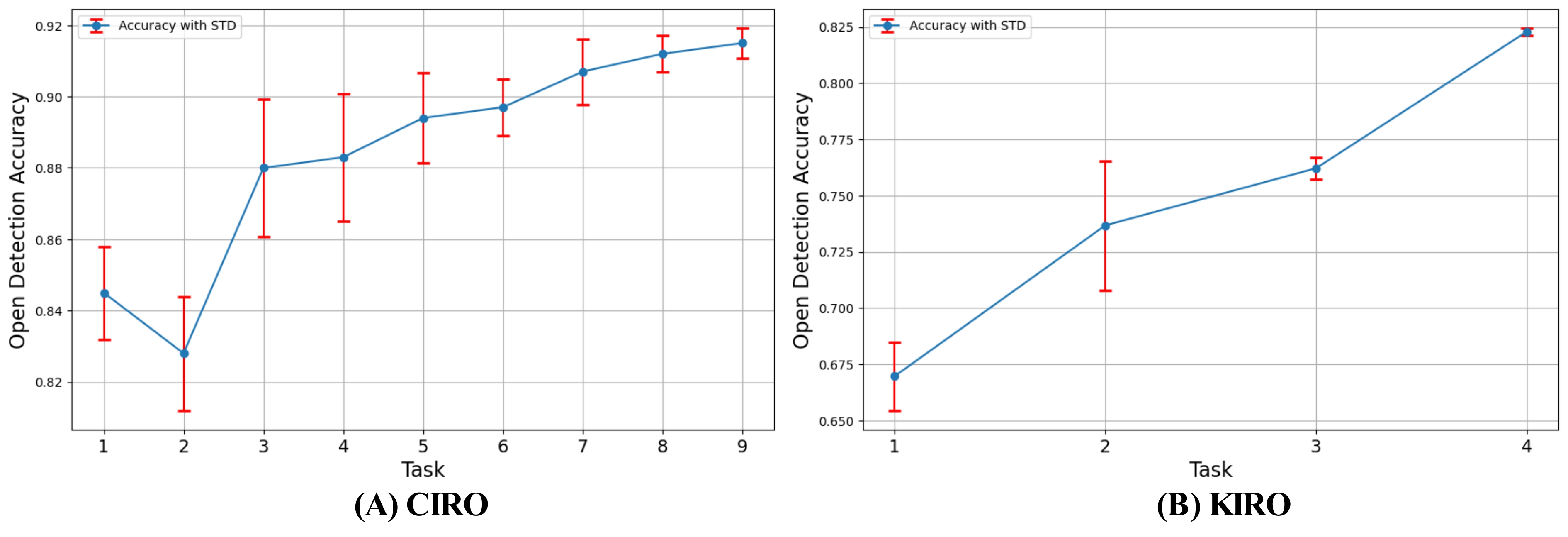}
    \caption{\textbf{Visualizations of Main Results.} The experimental results indicate that, for both CIRO and KIRO scenarios, the accuracy of open set detection shows an increasing trend as the incremental process progresses with new tasks appearing. The standard deviation of accuracy across different random seeds significantly decreases, suggesting that the proposed HoliTrans can incrementally learn the changing distributions of training data from new tasks, thereby accumulating knowledge, reducing open set risk, and thus demonstrating robustness.}
    \label{fig:enter-label}
\end{figure*}

\begin{table*}[ht]
\centering
\caption{Ablation Study.}\label{tab:ablation}
\vspace{2mm}
\begin{tabular}{ccccccc}
\toprule
\multirow{2}{*}{\textit{\textbf{Ablation Component}}}                   & \multicolumn{3}{c}{\textbf{Scenario: CIRO (Split-CIFAR100)}}                                           & \multicolumn{3}{c}{\textbf{Scenario: KIRO (Open-CORe50)}}                                           \\ \cmidrule(lr){2-4}\cmidrule(lr){5-7}
                                                              & \multicolumn{1}{c}{$ACC_t(\uparrow)$} & \multicolumn{1}{c}{$AUC_t (\uparrow)$} & \multicolumn{1}{c}{$FPR_t (\downarrow)$} & \multicolumn{1}{c}{$ACC_t(\uparrow)$} & \multicolumn{1}{c}{$AUC_t (\uparrow)$} & \multicolumn{1}{c}{$FPR_t (\downarrow)$} \\ \hline
\begin{tabular}[c]{@{}c@{}}w/o ternary search\end{tabular} & $0.842_{\pm 0.010}$     & $0.850_{\pm 0.012}$     & $0.202_{\pm 0.030}$     & $0.609_{\pm 0.020}$     & $0.609_{\pm 0.020}$     & $0.055_{\pm 0.029}$     \\ 
w/o DAPs                                                       & $0.468_{\pm 0.042}$     & $0.092_{\pm 0.010}$     & $0.878_{\pm 0.039}$     & $0.472_{\pm 0.022}$     & $0.212_{\pm 0.017}$     & $0.963_{\pm 0.015} $   \\  
\textbf{HoliTrans}                                                           & $\boldsymbol{0.851_{\pm 0.010}}$     & $\boldsymbol{0.859_{\pm 0.012}}$     & $\boldsymbol{0.193_{\pm 0.024}}$     & $\boldsymbol{0.748_{\pm 0.010}}$     & $\boldsymbol{0.748_{\pm 0.010}}$     & $\boldsymbol{0.054_{\pm 0.031}}$     \\ \bottomrule
\end{tabular}
\end{table*}
\subsubsection{Main Results}
Our evaluation approach distinguishes itself from prior works by addressing open-set detection and the classification of known samples simultaneously, rather than treating these tasks separately with distinct metrics. To ensure fairness and robust assessment, we utilize three random seeds with shuffled task orders, averaging the results and reporting the standard deviations. \autoref{tab:main results} presents the results on the CIRO and KIRO benchmark scenarios, where the proposed method is comprehensively compared against 22 state-of-the-art approaches.

Firstly, a comparison of the first 20 rows and the last 3 rows reveals that simply combining continual learning (CL) and out-of-distribution (OOD) detection methods fails to achieve satisfactory performance under open-world continual learning (OWCL) settings. This underscores the necessity of addressing OWCL as a unified problem, a central premise of our approach. 
Secondly, our HoliTrans method significantly outperforms all baseline models in both $ACC_t$ and $AUC_t$, achieving a remarkable 7\% improvement over the best existing methods. This establishes HoliTrans's exceptional capability to enhance incremental classification accuracy and quality across diverse tasks.
Thirdly, the proposed HoliTrans demonstrates outstanding performance in $FPR_t$, highlighting its superior ability to mitigate open-set risks—a critical challenge in OWCL problems. This performance advantage ensures that the model remains robust when encountering unseen or anomalous data.

In addition, across all evaluated metrics, HoliTrans consistently surpasses all baselines, setting new benchmarks for performance in all scenarios. 
Beyond algorithmic contributions, this work provides significant advancements for the OWCL research community. 
Also, by integrating existing CL models with diverse OOD algorithms, we create a comprehensive and reproducible benchmark, fostering further progress in this field. The reproducibility of our code and methods ensures that future research can build upon our work with confidence.

Moreover, the practical implications of this study are noteworthy. Our methodology bridges the gap between CL and OOD techniques, equipping researchers and practitioners with an effective and unified framework to tackle real-world OWCL challenges. This contribution not only advances the state-of-the-art but also serves as a foundational resource for the OWCL community, promoting collaboration and innovation.

\subsubsection{Ablation Study}
To underscore the significance of the key components in the HoliTrans framework, we conducted comprehensive ablation studies, focusing on the impact of adaptively learning the optimal \(r\) through ternary search and the contribution of all DAPs. The results are detailed in \autoref{tab:ablation}.

First, we examined the effect of replacing the ternary search strategy for adaptively determining \(r\) with the traditional equal interval search approach. As shown in \autoref{tab:ablation}, this substitution led to a noticeable decline in performance across all evaluation metrics, with the most pronounced degradation observed in the challenging KIRO scenario. This outcome highlights the importance of dynamic learning \(r\), especially in scenarios characterized by complex and evolving distributional shifts across tasks.

Second, we assessed the contribution of DAPs by entirely removing them, thereby reverting the model to a baseline OWCL framework employing a MaxLogits-like strategy. The absence of DAPs resulted in a marked reduction in performance, emphasizing their critical role in mitigating open risk and enhancing the quality of incremental classification. The inclusion of DAPs enables the model to effectively address open-set challenges and maintain robustness in diverse and incrementally evolving environments, as evidenced by the substantial performance gains reported in \autoref{tab:ablation}.

These findings collectively demonstrate that both the adaptive learning of \(r\) and the integration of DAPs are indispensable for achieving superior performance in open-world continual learning tasks.

\begin{figure*}[ht]
    \centering
    \includegraphics[width=0.95\textwidth]{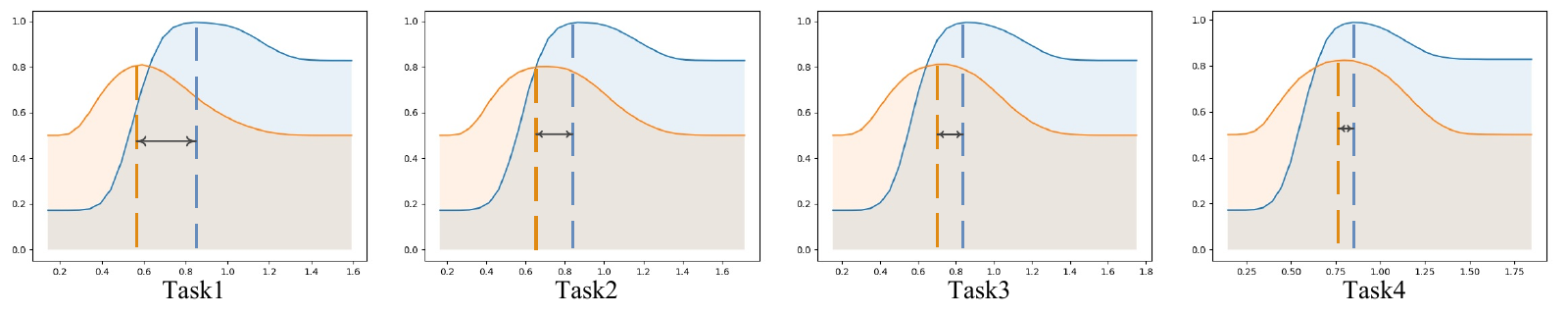}
    \caption{The divergence between the actual embedding distributions and the HoliTrans's embedding distributions with DAPs (The horizontal axis is the values of $r$, the vertical axis is $ACC_t$, and the dashed line indicates the optimal value of $r$). \textcolor{orange}{Orange} represents the actual ones, \textcolor{cyan}{blue} represents ours.}\label{fig:distribution}
\end{figure*}

\begin{table}[ht]\centering
\caption{Performance on Different NRP Size for \textbf{CIRO (Split-CIFAR100)} and \textbf{KIRO (Open-CORe50)}.}\label{table:combined}
\vspace{2mm}
\begin{tabular}{c|ccc}
\toprule
\multicolumn{4}{c}{\textbf{CIRO (Split-CIFAR100)}} \\ 
\hline
\textit{\textbf{Value of M}} & $ACC_t(\uparrow)$ & $AUC_t (\uparrow)$ & $FPR_t (\downarrow)$ \\ 
\hline
\textit{10000}                        & 0.851             & 0.825              & 0.142                \\
\textit{5000}                         & 0.859             & 0.825              & 0.137                \\
\textit{2500}                         & 0.863             & 0.877              & 0.124                \\
\textit{1250}                         & 0.856             & 0.876              & 0.104                \\
\textit{800}                          & 0.852             & 0.873              & 0.085                \\
\textit{400}                          & 0.833             & 0.860              & 0.059                \\ 
\toprule
\multicolumn{4}{c}{\textbf{KIRO (Open-CORe50)}} \\ 
\hline
\textit{\textbf{Value of M}} & $ACC_t(\uparrow)$ & $AUC_t (\uparrow)$ & $FPR_t (\downarrow)$ \\ 
\hline
\textit{10000}                        & 0.748             & 0.748              & 0.097                \\
\textit{5000}                         & 0.754             & 0.753              & 0.055                \\
\textit{2500}                         & 0.744             & 0.744              & 0.063                \\
\textit{1250}                         & 0.744             & 0.744              & 0.051                \\
\textit{800}                          & 0.737             & 0.737              & 0.041                \\
\textit{400}                          & 0.756             & 0.756              & 0.048                \\ 
\toprule
\end{tabular}
\end{table}
\subsubsection{Generalization and Robustness}
\begin{figure}
    \centering
    \includegraphics[width=\linewidth]{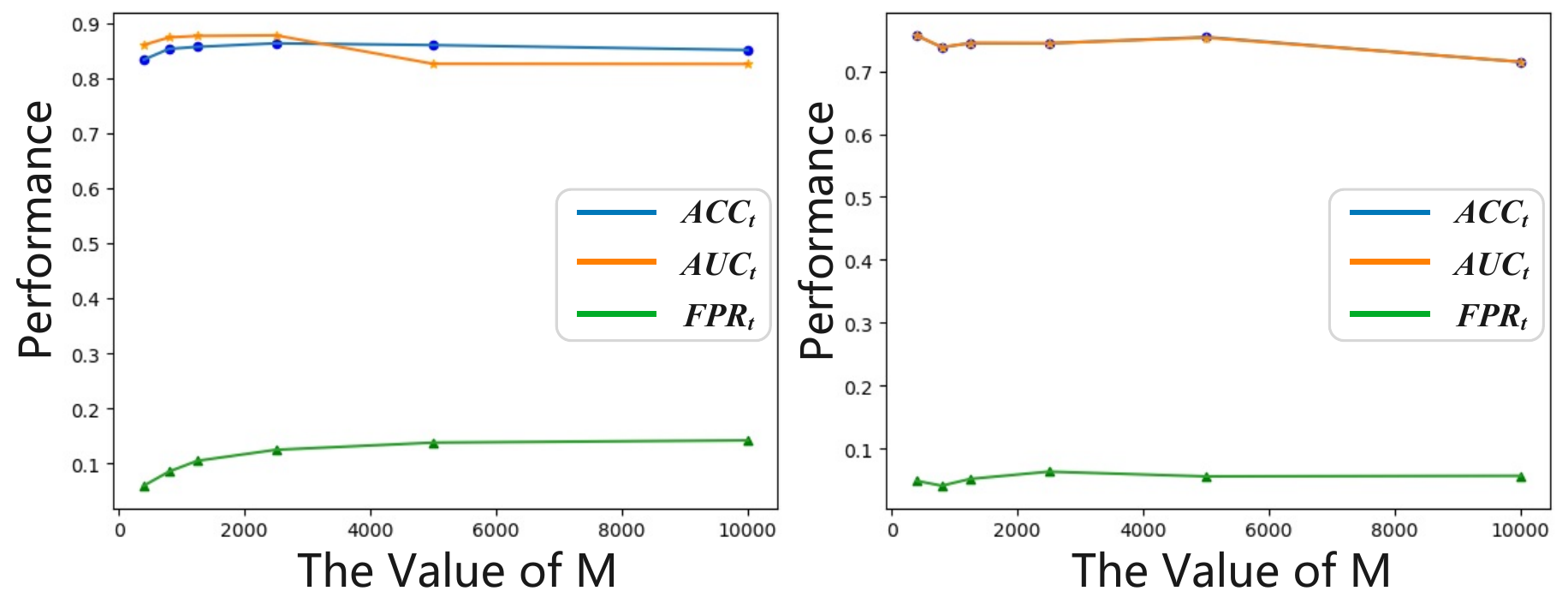}
    \caption{Performance vs. NRP Size.}
    \label{fig2}
\end{figure}

We evaluate the impact of the size of NRP (i.e., M) and report the final averaged performance across all tasks on different scenarios.
As shown in \autoref{fig2} and \autoref{table:combined}, our HoliTrans remains robust across different values of M from 400 to 10000. Notably, a larger M for the random projection size does not necessarily lead to better performance: HoliTrans achieves better performance when M is between 1000 and 2000, which is a relatively small dimension compared to previous works and shows the adaptability of NPR on the OWCL problems.

Additionally, we compare the actual embedding distribution with the embedding distribution learned by HoliTrans. As shown in \autoref{fig:distribution}, HoliTrans increasingly aligns with the actual embedding distribution as new tasks come, with decreasing deviation.
Such results highlight the potential of HoliTrans to adapt to new tasks and accumulate knowledge.
Furthermore, to verify the impact of experience buffer $\mathcal{B}$, we replace our DAPs with a KNN-contrastive method \cite{zhou2022knn} that replays embeddings in different sizes. The results illustrate that our HoliTrans with DAPs uses less memory and performs better in all OWCL scenarios.

\section{Conclusions and Limitation Analysis}
Despite the increasing interest in OWCL research, existing approaches merely focused on learning and transferring knowledge from known samples, while overlooking the knowledge derived from unknown samples. In this paper, our empirical findings not only challenge the current assumption, but also underscore the critical importance of addressing the OWCL problem as a holistic paradigm. Following this, we introduce a novel OWCL approach, HoliTrans, that supports the knowns-unknowns knowledge transfer and validates the proposed model through comprehensive experiments.

The key insight of our work lies in discovering the interaction between open risk and incremental prediction error, viewing OWCL as a holistic problem rather than a mere combination of continual learning and open-set recognition methods. Additionally, we cast OWCL into 4 scenarios with increasing difficulty and conduct extensive experiments to investigate the challenges of OWCL. More importantly, our theoretical analysis provides guidance for proposing a principled and novel OWCL framework. Specifically, we introduce the nonlinear random projection (NRP) method, which leverages nonlinear random feature interactions to enhance linear separability in a higher-dimensional space and propose distribution-aware prototypes (DAPs) for seen classes to mitigate open risk and enhance incremental classification performance. Experimental results demonstrate the superiority of HoliTrans over 22 baseline models across various OWCL scenarios and datasets, further underscoring its capability to effectively learn and transfer knowledge from both known and unknown samples.

While the theoretical advancements and extensive empirical validation underscore the superiority of our HoliTrans, certain limitations warrant consideration: 1) The practical implementation of theoretical results may encounter challenges due to the few-shot training and noise in real-world data, potentially affecting the algorithm's performance. 2) The generalizability of the experimental findings might be constrained by task similarity, necessitating further validation across diverse and complex situations that consider task similarity. 3) Experimental results show that using prototypes with lower memory usage can achieve better performance than replaying a large number of samples in experience buffers. This phenomenon also warrants deeper theoretical investigation. Hence, we will elaborate on these points in our future work.

\section*{Acknowledgment}
This work was supported by the National Natural Science Foundation of China (Nos. 62476228, 62406259), the Natural Science Foundation of Sichuan Province (No. 2022NSFSC0528), the Sichuan Science and Technology Program (No. 2024ZYD0180) and the Fundamental Research Funds for the Central Universities (YJ202421).

\ifCLASSOPTIONcaptionsoff
  \newpage
\fi

\bibliographystyle{IEEEtran}
\bibliography{main.bib}

\begin{thebibliography}{10}
\providecommand{\url}[1]{#1}
\csname url@samestyle\endcsname
\providecommand{\newblock}{\relax}
\providecommand{\bibinfo}[2]{#2}
\providecommand{\BIBentrySTDinterwordspacing}{\spaceskip=0pt\relax}
\providecommand{\BIBentryALTinterwordstretchfactor}{4}
\providecommand{\BIBentryALTinterwordspacing}{\spaceskip=\fontdimen2\font plus
\BIBentryALTinterwordstretchfactor\fontdimen3\font minus \fontdimen4\font\relax}
\providecommand{\BIBforeignlanguage}[2]{{%
\expandafter\ifx\csname l@#1\endcsname\relax
\typeout{** WARNING: IEEEtran.bst: No hyphenation pattern has been}%
\typeout{** loaded for the language `#1'. Using the pattern for}%
\typeout{** the default language instead.}%
\else
\language=\csname l@#1\endcsname
\fi
#2}}
\providecommand{\BIBdecl}{\relax}
\BIBdecl

\bibitem{kim2022theoretical}
G.~Kim, C.~Xiao, T.~Konishi, Z.~Ke, and B.~Liu, ``A theoretical study on solving continual learning,'' \emph{Advances in neural information processing systems}, vol.~35, pp. 5065--5079, 2022.

\bibitem{liu2023ai}
B.~Liu, S.~Mazumder, E.~Robertson, and S.~Grigsby, ``Ai autonomy: Self-initiation, adaptation and continual learning,'' \emph{AI Magazine}, 2023.

\bibitem{scheirer2012toward}
W.~J. Scheirer, A.~de~Rezende~Rocha, A.~Sapkota, and T.~E. Boult, ``Toward open set recognition,'' \emph{IEEE transactions on pattern analysis and machine intelligence}, vol.~35, no.~7, pp. 1757--1772, 2012.

\bibitem{bendale2015towards}
A.~Bendale and T.~Boult, ``Towards open world recognition,'' in \emph{Proceedings of the IEEE Conference on Computer Vision and Pattern Recognition}, 2015, pp. 1893--1902.

\bibitem{kirkpatrick2017overcoming}
J.~Kirkpatrick, R.~Pascanu, N.~Rabinowitz, J.~Veness, G.~Desjardins, A.~A. Rusu, K.~Milan, J.~Quan, T.~Ramalho, A.~Grabska-Barwinska \emph{et~al.}, ``Overcoming catastrophic forgetting in neural networks,'' \emph{Proceedings of the national academy of sciences}, vol. 114, no.~13, pp. 3521--3526, 2017.

\bibitem{pmlr-v70-zenke17a}
F.~Zenke, B.~Poole, and S.~Ganguli, ``Continual learning through synaptic intelligence,'' in \emph{Proceedings of the 34th International Conference on Machine Learning}, ser. Proceedings of Machine Learning Research, D.~Precup and Y.~W. Teh, Eds., vol.~70.\hskip 1em plus 0.5em minus 0.4em\relax PMLR, 06--11 Aug 2017, pp. 3987--3995.

\bibitem{wang2024comprehensive}
L.~Wang, X.~Zhang, H.~Su, and J.~Zhu, ``A comprehensive survey of continual learning: theory, method and application,'' \emph{IEEE Transactions on Pattern Analysis and Machine Intelligence}, 2024.

\bibitem{kishida2021object}
I.~Kishida, H.~Chen, M.~Baba, J.~Jin, A.~Amma, and H.~Nakayama, ``Object recognition with continual open set domain adaptation for home robot,'' in \emph{Proceedings of the IEEE/CVF Winter Conference on Applications of Computer Vision}, 2021, pp. 1517--1526.

\bibitem{truong2023fairness}
T.-D. Truong, H.-Q. Nguyen, B.~Raj, and K.~Luu, ``Fairness continual learning approach to semantic scene understanding in open-world environments,'' \emph{Advances in Neural Information Processing Systems}, vol.~36, pp. 65\,456--65\,467, 2023.

\bibitem{kim2025open}
G.~Kim, C.~Xiao, T.~Konishi, Z.~Ke, and B.~Liu, ``Open-world continual learning: Unifying novelty detection and continual learning,'' \emph{Artificial Intelligence}, vol. 338, p. 104237, 2025.

\bibitem{zhou2024class}
D.-W. Zhou, Q.-W. Wang, Z.-H. Qi, H.-J. Ye, D.-C. Zhan, and Z.~Liu, ``Class-incremental learning: A survey,'' \emph{IEEE Transactions on Pattern Analysis and Machine Intelligence}, 2024.

\bibitem{parisi2019continual}
G.~I. Parisi, J.~Tani, C.~Weber, and S.~Wermter, ``Continual learning with dual memory,'' \emph{IEEE Transactions on Neural Networks and Learning Systems}, vol.~30, no.~4, pp. 1053--1065, 2019.

\bibitem{dhar2019learning}
P.~Dhar, R.~V. Singh, K.-C. Peng, Z.~Wu, and R.~Chellappa, ``Learning without memorizing,'' in \emph{Proceedings of the IEEE/CVF Conference on Computer Vision and Pattern Recognition (CVPR)}.\hskip 1em plus 0.5em minus 0.4em\relax IEEE, 2019, pp. 5138--5146.

\bibitem{vijayan2023trire}
P.~Vijayan, P.~Bhat, B.~Zonooz, and E.~Arani, ``Trire: A multi-mechanism learning paradigm for continual knowledge retention and promotion,'' in \emph{Advances in Neural Information Processing Systems}, 2023.

\bibitem{mundt2022unified}
M.~Mundt, I.~Pliushch, S.~Majumder, Y.~Hong, and V.~Ramesh, ``Unified probabilistic deep continual learning through generative replay and open set recognition,'' \emph{Journal of Imaging}, vol.~8, no.~4, p.~93, 2022.

\bibitem{masana2022class}
M.~Masana, X.~Liu, B.~Twardowski, M.~Menta, A.~D. Bagdanov, and J.~Van De~Weijer, ``Class-incremental learning: survey and performance evaluation on image classification,'' \emph{IEEE Transactions on Pattern Analysis and Machine Intelligence}, vol.~45, no.~5, pp. 5513--5533, 2022.

\bibitem{lo2022adversarially}
S.-Y. Lo, P.~Oza, and V.~M. Patel, ``Adversarially robust one-class novelty detection,'' \emph{IEEE Transactions on Pattern Analysis and Machine Intelligence}, vol.~45, no.~4, pp. 4167--4179, 2022.

\bibitem{li2024learning}
Y.~Li, X.~Yang, H.~Wang, X.~Wang, and T.~Li, ``Learning to prompt knowledge transfer for open-world continual learning,'' in \emph{Proceedings of the AAAI Conference on Artificial Intelligence}, vol. 38(12), 2024, pp. 13\,700--13\,708.

\bibitem{schlachter2020deep}
P.~Schlachter, Y.~Liao, and B.~Yang, ``Deep open set recognition using dynamic intra-class splitting,'' \emph{SN Computer Science}, vol.~1, pp. 1--12, 2020.

\bibitem{zhu2022multi}
P.~Zhu, W.~Zhang, Y.~Wang, and Q.~Hu, ``Multi-granularity inter-class correlation based contrastive learning for open set recognition.'' \emph{Int. J. Softw. Informatics}, vol.~12, no.~2, pp. 157--175, 2022.

\bibitem{huang2022class}
H.~Huang, Y.~Wang, Q.~Hu, and M.-M. Cheng, ``Class-specific semantic reconstruction for open set recognition,'' \emph{IEEE transactions on pattern analysis and machine intelligence}, vol.~45, no.~4, pp. 4214--4228, 2022.

\bibitem{zhou2022knn}
Y.~Zhou, P.~Liu, and X.~Qiu, ``Knn-contrastive learning for out-of-domain intent classification,'' in \emph{Proceedings of the 60th Annual Meeting of the Association for Computational Linguistics (Volume 1: Long Papers)}, 2022, pp. 5129--5141.

\bibitem{li2017learning}
Z.~Li and D.~Hoiem, ``Learning without forgetting,'' \emph{IEEE transactions on pattern analysis and machine intelligence}, vol.~40, no.~12, pp. 2935--2947, 2017.

\bibitem{fei2016learning}
G.~Fei, S.~Wang, and B.~Liu, ``Learning cumulatively to become more knowledgeable,'' in \emph{Proceedings of the 22nd ACM SIGKDD International Conference on Knowledge Discovery and Data Mining}, 2016, pp. 1565--1574.

\bibitem{liu2024task}
X.~Liu, J.-T. Zhai, A.~D. Bagdanov, K.~Li, and M.-M. Cheng, ``Task-adaptive saliency guidance for exemplar-free class incremental learning,'' in \emph{Proceedings of the IEEE/CVF Conference on Computer Vision and Pattern Recognition}, 2024, pp. 23\,954--23\,963.

\bibitem{wang2020novelty}
Y.~Wang, Y.~Ding, X.~He, X.~Fan, C.~Lin, F.~Li, T.~Wang, Z.~Luo, and J.~Luo, ``Novelty detection and online learning for chunk data streams,'' \emph{IEEE Transactions on Pattern Analysis and Machine Intelligence}, vol.~43, no.~7, pp. 2400--2412, 2020.

\bibitem{yu2022self}
L.~Yu, X.~Liu, and J.~Van~de Weijer, ``Self-training for class-incremental semantic segmentation,'' \emph{IEEE Transactions on Neural Networks and Learning Systems}, vol.~34, no.~11, pp. 9116--9127, 2022.

\bibitem{aljundi2022continual}
R.~Aljundi, D.~O. Reino, N.~Chumerin, and R.~E. Turner, ``Continual novelty detection,'' in \emph{Conference on Lifelong Learning Agents}.\hskip 1em plus 0.5em minus 0.4em\relax PMLR, 2022, pp. 1004--1025.

\bibitem{8631004-2019}
S.~Dang, Z.~Cao, Z.~Cui, Y.~Pi, and N.~Liu, ``Open set incremental learning for automatic target recognition,'' \emph{IEEE Transactions on Geoscience and Remote Sensing}, vol.~57, no.~7, pp. 4445--4456, 2019.

\bibitem{Joseph_2021_CVPR}
K.~J. Joseph, S.~Khan, F.~S. Khan, and V.~N. Balasubramanian, ``Towards open world object detection,'' in \emph{Proceedings of the IEEE/CVF Conference on Computer Vision and Pattern Recognition}, June 2021, pp. 5830--5840.

\bibitem{chan2021entropy}
R.~Chan, M.~Rottmann, and H.~Gottschalk, ``Entropy maximization and meta classification for out-of-distribution detection in semantic segmentation,'' in \emph{Proceedings of the ieee/cvf international conference on computer vision}, 2021, pp. 5128--5137.

\bibitem{mazumder2024lifelong}
S.~Mazumder and B.~Liu, \emph{Lifelong and Continual Learning Dialogue Systems}.\hskip 1em plus 0.5em minus 0.4em\relax Springer Nature, 2024.

\bibitem{van2018three}
G.~M. van~de Ven and A.~S. Tolias, ``Three continual learning scenarios,'' in \emph{NeurIPS Continual Learning Workshop}, vol. 1(9), 2018.

\bibitem{mcdonnell2024ranpac}
M.~D. McDonnell, D.~Gong, A.~Parvaneh, E.~Abbasnejad, and A.~van~den Hengel, ``Ranpac: Random projections and pre-trained models for continual learning,'' \emph{Advances in Neural Information Processing Systems}, vol.~36, 2024.

\bibitem{wang2022learning}
Z.~Wang, Z.~Zhang, C.-Y. Lee, H.~Zhang, R.~Sun, X.~Ren, G.~Su, V.~Perot, J.~Dy, and T.~Pfister, ``Learning to prompt for continual learning,'' in \emph{Proceedings of the IEEE/CVF Conference on Computer Vision and Pattern Recognition}, 2022, pp. 139--149.

\bibitem{zhou2023revisiting}
D.-W. Zhou, H.-J. Ye, D.-C. Zhan, and Z.~Liu, ``Revisiting class-incremental learning with pre-trained models: Generalizability and adaptivity are all you need,'' \emph{arXiv preprint arXiv:2303.07338}, 2023.

\bibitem{fang2021learning}
Z.~Fang, J.~Lu, A.~Liu, F.~Liu, and G.~Zhang, ``Learning bounds for open-set learning,'' in \emph{International conference on machine learning}.\hskip 1em plus 0.5em minus 0.4em\relax PMLR, 2021, pp. 3122--3132.

\bibitem{shi2024unified}
H.~Shi and H.~Wang, ``A unified approach to domain incremental learning with memory: Theory and algorithm,'' \emph{Advances in Neural Information Processing Systems}, vol.~36, 2024.

\bibitem{vapnik1998statistical}
V.~N. Vapnik, V.~Vapnik \emph{et~al.}, ``Statistical learning theory,'' \emph{Wiley New York}, 1998.

\bibitem{anthony1999neural}
M.~Anthony, P.~L. Bartlett, P.~L. Bartlett \emph{et~al.}, \emph{Neural network learning: Theoretical foundations}.\hskip 1em plus 0.5em minus 0.4em\relax cambridge university press Cambridge, 1999, vol.~9.

\bibitem{houlsby2019parameter}
N.~Houlsby, A.~Giurgiu, S.~Jastrzebski, B.~Morrone, Q.~de~Laroussilhe, A.~Gesmundo, M.~Attariyan, and S.~Gelly, ``Parameter-efficient transfer learning for {NLP},'' in \emph{Proceedings of the 36th International Conference on Machine Learning (ICML 2019)}.\hskip 1em plus 0.5em minus 0.4em\relax PMLR, 2019, pp. 2790--2799.

\bibitem{li2021prefix}
X.~Li, Y.~Liang, Z.~Ma, Y.~Li, and S.~C.~H. Hoi, ``Prefix-tuning: Optimizing continuous prompts for generation,'' in \emph{Association for Computational Linguistics (ACL) 2021}, 2021.

\bibitem{chen2022adaptformer}
S.~Chen, C.~Ge, Z.~Tong, J.~Wang, Y.~Song, J.~Wang, and P.~Luo, ``Adaptformer: Adapting vision transformers for scalable visual recognition,'' \emph{Advances in Neural Information Processing Systems}, vol.~35, pp. 16\,664--16\,678, 2022.

\bibitem{lian2022scaling}
D.~Lian, D.~Zhou, J.~Feng, and X.~Wang, ``Scaling \& shifting your features: A new baseline for efficient model tuning,'' \emph{Advances in Neural Information Processing Systems}, vol.~35, pp. 109--123, 2022.

\bibitem{jia2022visual}
M.~Jia, L.~Tang, B.-C. Chen, C.~Cardie, S.~Belongie, B.~Hariharan, and S.-N. Lim, ``Visual prompt tuning,'' in \emph{European Conference on Computer Vision}.\hskip 1em plus 0.5em minus 0.4em\relax Springer, 2022, pp. 709--727.

\bibitem{chen1996rapid}
C.~P. Chen, ``A rapid supervised learning neural network for function interpolation and approximation,'' \emph{IEEE Transactions on Neural Networks}, vol.~7, no.~5, pp. 1220--1230, 1996.

\bibitem{huang2006extreme}
G.-B. Huang, Q.-Y. Zhu, and C.-K. Siew, ``Extreme learning machine: theory and applications,'' \emph{Neurocomputing}, vol.~70, no. 1-3, pp. 489--501, 2006.

\bibitem{wang2022dualprompt}
Z.~Wang, Z.~Zhang, S.~Ebrahimi, R.~Sun, H.~Zhang, C.-Y. Lee, X.~Ren, G.~Su, V.~Perot, J.~Dy \emph{et~al.}, ``Dualprompt: Complementary prompting for rehearsal-free continual learning,'' in \emph{European Conference on Computer Vision}.\hskip 1em plus 0.5em minus 0.4em\relax Springer, 2022, pp. 631--648.

\bibitem{smith2023coda}
J.~S. Smith, L.~Karlinsky, V.~Gutta, P.~Cascante-Bonilla, D.~Kim, A.~Arbelle, R.~Panda, R.~Feris, and Z.~Kira, ``Coda-prompt: Continual decomposed attention-based prompting for rehearsal-free continual learning,'' in \emph{Proceedings of the IEEE/CVF Conference on Computer Vision and Pattern Recognition}, 2023, pp. 11\,909--11\,919.

\bibitem{basart2022scaling}
S.~Basart, M.~Mantas, M.~Mohammadreza, S.~Jacob, and S.~Dawn, ``Scaling out-of-distribution detection for real-world settings,'' in \emph{International Conference on Machine Learning}, 2022.

\bibitem{liu2020energy}
W.~Liu, X.~Wang, J.~Owens, and Y.~Li, ``Energy-based out-of-distribution detection,'' \emph{Advances in Neural Information Processing Systems}, vol.~33, pp. 21\,464--21\,475, 2020.

\bibitem{krizhevsky2009learning}
A.~Krizhevsky, G.~Hinton \emph{et~al.}, ``Learning multiple layers of features from tiny images,'' \emph{Master's thesis, University of Tront}, 2009.

\bibitem{lomonaco2017core50}
V.~Lomonaco and D.~Maltoni, ``Core50: a new dataset and benchmark for continuous object recognition,'' in \emph{Conference on robot learning}.\hskip 1em plus 0.5em minus 0.4em\relax PMLR, 2017, pp. 17--26.

\end{thebibliography}

\vfill

\end{document}